%% file: egpaper_final.tex
\documentclass[10pt,twocolumn,letterpaper]{article}

\usepackage{iccv}
\usepackage{times}
\usepackage{epsfig}
\usepackage{graphicx}
\usepackage{amsmath}
\usepackage{amssymb}

\usepackage[font=small,skip=6pt]{caption}
\usepackage[labelformat=simple,font={small}]{subcaption}
\usepackage[flushleft]{threeparttable}

\usepackage{booktabs}
\usepackage{multirow}

\usepackage[dvipsnames]{xcolor}
\usepackage{color, colortbl}

\usepackage[breaklinks=true,bookmarks=false]{hyperref}

\usepackage[capitalize]{cleveref}
\crefname{section}{Sec.}{Secs.}
\Crefname{section}{Section}{Sections}
\Crefname{table}{Table}{Tables}
\crefname{table}{Tab.}{Tabs.}

\iccvfinalcopy 

\def\iccvPaperID{3721} 

\ificcvfinal\fi

\begin{document}

\definecolor{ForestGreen}{rgb}{0.13, 0.55, 0.13}
\definecolor{ForestGreen2}{rgb}{0.2, 0.5372549019607843, 0.1803921568627451}
\definecolor{tsne_yellow}{RGB}{253, 231, 37}
\definecolor{tsne_purple}{RGB}{68, 1, 83}
\newcommand{\FGT}[1]{\textcolor{ForestGreen2}{#1}}
\newcommand{\FG}[1]{\textcolor{ForestGreen}{\textbf{#1}}}
\newcommand{\Gd}{\rowcolor{gray!45}}

\setlength{\textfloatsep}{10pt}
\setlength{\intextsep}{10pt}
\setlength{\floatsep}{10pt}
\setlength{\dbltextfloatsep}{10pt}
\setlength{\dblfloatsep}{10pt}

\title{Confidence-based Visual Dispersal for Few-shot \\ Unsupervised Domain Adaptation}

\author{Yizhe Xiong$^{1,2,3}$\quad Hui Chen$^{2}$\thanks{Corresponding author.}\quad Zijia Lin$^{1}$\quad Sicheng Zhao$^{2}$\quad Guiguang Ding$^{1,2}$\footnotemark[1]\\
$^{1}$School of Software, Tsinghua University\\ 
$^{2}$Beijing National Research Center for Information Science and Technology (BNRist)\\
$^{3}$Hangzhou Zhuoxi Institute of Brain and Intelligence \\
{\tt\small \{xiongyizhe2001, jichenhui2012, schzhao\}@gmail.com}\\
{\tt\small linzijia07@tsinghua.org.cn dinggg@tsinghua.edu.cn}
}

\maketitle
\ificcvfinal\thispagestyle{empty}\fi

\begin{abstract}
   Unsupervised domain adaptation aims to transfer knowledge from a fully-labeled source domain to an unlabeled target domain. 
   However, in real-world scenarios, providing abundant labeled data even in the source domain can be infeasible due to the difficulty and high expense of annotation.
   To address this issue, recent works consider the Few-shot Unsupervised Domain Adaptation (FUDA) where only a few source samples are labeled, and conduct knowledge transfer via self-supervised learning methods.
   Yet existing methods generally overlook that the sparse label setting hinders learning reliable source knowledge for transfer.
   Additionally, the learning difficulty difference in target samples is different but ignored, leaving hard target samples poorly classified.
   To tackle both deficiencies, in this paper, we propose a novel Confidence-based Visual Dispersal Transfer learning method (C-VisDiT) for FUDA.
   Specifically, C-VisDiT consists of a cross-domain visual dispersal strategy that transfers only high-confidence source knowledge for model adaptation and an intra-domain visual dispersal strategy that guides the learning of hard target samples with easy ones.
   We conduct extensive experiments on Office-31, Office-Home, VisDA-C, and DomainNet benchmark datasets and the results demonstrate that the proposed C-VisDiT significantly outperforms state-of-the-art FUDA methods. 
   Our code is available at \url{https://github.com/Bostoncake/C-VisDiT}.
\end{abstract}

\section{Introduction}
\label{sec:intro}

Deep learning has achieved considerably high performance in various computer vision tasks, such as instance recognition~\cite{DBLP:conf/cvpr/HeZRS16,DBLP:conf/eccv/ZeilerF14}, semantic segmentation~\cite{DBLP:conf/cvpr/LongSD15,DBLP:journals/corr/LiuRB15,DBLP:journals/pami/BadrinarayananK17}, and object detection~\cite{DBLP:conf/nips/RenHGS15,DBLP:conf/eccv/CarionMSUKZ20,DBLP:conf/iccv/LiuL00W0LG21}.
Despite the remarkable success, well-tuned deep models usually suffer from dramatic performance drops when being applied in real-world scenarios because of the domain gap issue~\cite{torralba2011unbiased,tzeng2017adversarial,zhao2022review,wilson2020survey}, \ie the shift between the learnt distribution and the testing distribution.
To cope with such performance degradation, researchers have been dedicated to unsupervised domain adaptation (UDA)~\cite{DBLP:conf/cvpr/LongSD15,DBLP:journals/jmlr/GaninUAGLLML16,tzeng2017adversarial,long2018conditional} aiming at transferring the learned knowledge from a labeled source domain to another unlabeled target domain. 

Though being promising, UDA assumes that a fully-annotated source domain can be conveniently accessed for training. However, in some real-world scenarios, such an assumption is violated, ending up with sparsely labeled source data.
For example, in the field of chip defect detection, it is unrealistic to perfectly annotate massive defect data in the source domain, due to the strict quality regulation and high expense of annotation~\cite{DBLP:journals/aei/ChenT21,huang2022small}.
In the medical industry, the high cost and difficulty of labeling also prohibit providing sufficient labeled source data~\cite{DBLP:conf/cvpr/WangZZW20,liru2022pneumonia}. 
Therefore, how to realize UDA with sparsely labeled source data becomes appealing and has attracted increasing attention~\cite{DBLP:journals/corr/abs-2003-08264,DBLP:conf/cvpr/YueZZ0DKS21,abbet2021selfrule,rakshit2022open}.

Recently, researchers address this issue by considering the Few-shot Unsupervised Domain Adaptation (FUDA), where only a few source samples are labeled.
Compared with UDA, FUDA is more challenging due to the insufficiency of the source knowledge.
With only a few labeled source samples, deep models are difficult to learn a discriminative feature space to represent the complicated semantic information, thus largely affecting the knowledge transfer. 
And thus to compensate for the shortage of labeled source data, existing methods~\cite{DBLP:journals/corr/abs-2003-08264,DBLP:conf/cvpr/YueZZ0DKS21} generally employ self-supervised learning strategies to mine the latent correspondence between the source distribution and the target one. Specifically, Kim \etal~\cite{DBLP:journals/corr/abs-2003-08264} proposed a cross-domain self-supervised learning method (CDS) to derive features that are not only domain-invariant but also class-discriminative for the target domain. Yue \etal~\cite{DBLP:conf/cvpr/YueZZ0DKS21} further enhanced the adaptation with an end-to-end prototypical cross-domain self-supervised learning method (PCS) which achieves the state-of-the-art performance. 

Despite great progress, existing self-supervised FUDA methods transfer knowledge from all visible samples without considering the discrimination of their individual confidence towards model adaptation, which exhibits two inherent weaknesses. 
First, existing methods generally align target features with all source features including unlabeled ones.
Although having performance gains, such a strategy discards that the sparse label setting of FUDA inevitably introduces large noise, resulting in the confidence inconsistency among source samples and thus transfering unreliable knowledge.
Second, existing methods generally feed all target data into the self-supervised learning modules, assuming that the model optimizes equally on each target sample. However, difficult target samples, often presented as samples with low prediction confidence, are not well-learned during the adaptation. 

\begin{figure}[t]
  \centering
  \includegraphics[width=1.0\linewidth]{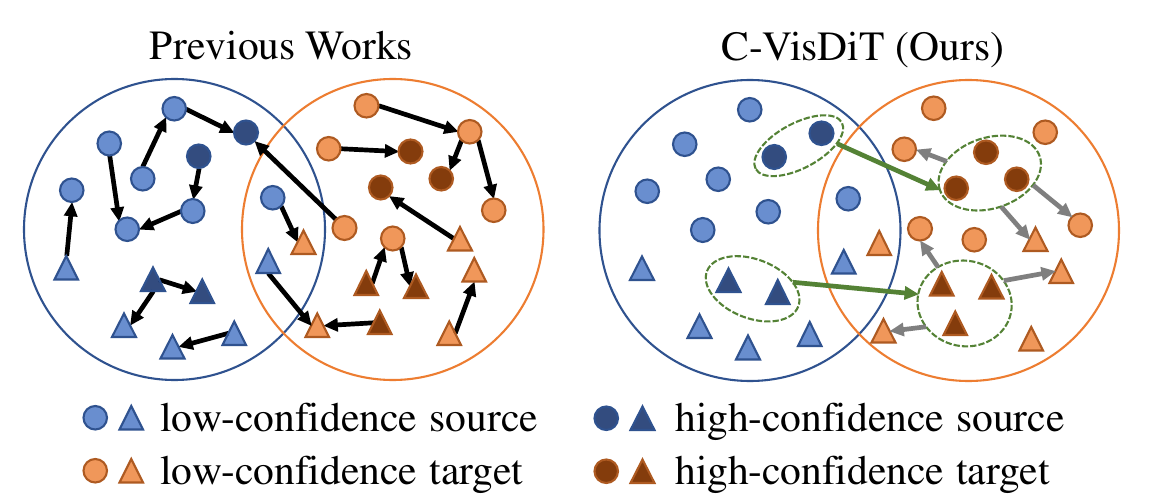}

   \caption{We address the task of few-shot unsupervised domain adaptation. Left: Existing FUDA methods utilize self-supervised learning methods to implicitly align source and target domains without considering the confidence of samples. Right: Our C-VisDiT transfers knowledge from high-confidence source samples and high-confidence target samples to enhance model adaptation. }
   \label{fig:figure_abs}
\end{figure}

To tackle those issues, in this paper, we propose a novel \textbf{C}onfidence-based \textbf{Vis}ual \textbf{Di}spersal \textbf{T}ransfer learning method (\textbf{C-VisDiT}) for FUDA, aiming to comprehensively consider the different importance of each sample based on its confidence, as illustrated in \cref{fig:figure_abs}. 
Considering the huge difficulty of learning a neat representation with very few labeled source samples, we propose to disperse the knowledge directly at the visual\footnote{Here we refer to ``visual'' as the raw RGB image information.} level. Specifically, for the source domain, we introduce a cross-domain visual dispersal to enhance the knowledge transfer from high-confidence source samples to unlabeled target ones. 
Without the disturbance of unreliable low-confidence source knowledge, our cross-domain visual dispersal can achieve more reliable cross-domain knowledge transfer.
Besides, we perform an intra-domain visual dispersal between samples of different learning difficulties for the target domain, aiming to guide the learning of hard target samples by the easy ones. In contrast to existing methods, our intra-domain visual dispersal can diminish the learning disparity between target samples of different learning difficulties, thus boosting the model adaptation.

We emphasize that the core contribution of our C-VisDiT is the confidence-based strategies, motivated by the idea that samples with low prediction confidence greatly harm the training in the label-scarse scenario of FUDA. Although we conduct visual dispersal via MixUp~\cite{DBLP:conf/iclr/ZhangCDL18}, 
we show that our C-VisDiT and its variants can greatly outperform their vanilla MixUp-based alternatives which completely discard the confidence of samples, well confirming the importance of the proposed confidence-based transfer learning for FUDA.

To summarize, our contributions are three-fold:

\begin{itemize}

    \item We propose a confidence-based visual dispersal transfer learning method (C-VisDiT) for FUDA, which simultaneously takes the reliability of the source samples and the learning difficulty of the target samples into consideration.
    
    \item We introduce a cross-domain visual dispersal to avoid the negative impact of source knowledge with low confidence. An intra-domain visual dispersal is also employed to boost the learning of hard unlabeled target samples with the guidance of easy ones.
    
    \item We conduct extensive experiments on four popular benchmark datasets for FUDA, \ie, Office-31~\cite{saenko2010adapting}, Office-home~\cite{venkateswara2017deep}, VisDA-C~\cite{peng2017visda}, and DomainNet~\cite{peng2019moment}. Experiment results show that our C-VisDiT consistently outperforms existing methods with a significant margin, 
    establishing new state-of-the-art results for all datasets. That well demonstrate the effectiveness and superiority of our method.

\end{itemize}
\section{Related Work}
\label{sec:related}

\textbf{Unsupervised Domain Adaptation (UDA).}
UDA aims to transfer the knowledge from a fully labeled source domain to an unlabeled target domain. 
Most UDA methods tend to learn domain-invariant feature representations. Earlier works in UDA aim to optimize feature extractors explicitly by domain divergence minimization \cite{gretton2012kernel,long2015learning,long2016unsupervised,zhuo2017deep,sun2019unsupervised}, adversarial domain classifier confusion~\cite{tzeng2017adversarial,hoffman2018cycada,long2018conditional}, or cross-domain data reconstruction~\cite{ghifary2016deep,bousmalis2016domain}. Recent works focus on the domain mapping~\cite{shrivastava2017learning,li2018semantic,bousmalis2018using}, which translates the source domain images to the target domain via conditional generative adversarial networks~\cite{zhu2017unpaired,yi2017dualgan,kim2017learning}.
Existing UDA methods utilize the abundant labeled source data, which is not well satisfied in some practical scenarios, like chip defect detection~\cite{DBLP:journals/aei/ChenT21} and medical image segmentation~\cite{perone2019unsupervised}.

Although designed for FUDA, our method can be adopted to the UDA problem. Empirical analysis show that the proposed confidence-based strategy can achieve slight improvements over baseline models. This is reasonable because the abundant source annotations can substantially guarantee the reliability of knowledge transfer, which, however, does not hold in FUDA. Therefore, we advocate thoughtfully managing each sample for both the source domain and the target domain, making our C-VisDiT possess a distinctive edge for FUDA.

\textbf{Few-shot Unsupervised Domain Adaptation (FUDA).}
FUDA transfers knowledge from a sparsely labeled source domain to an unlabeled target domain. 
Existing methods implicitly align the source and the target domains with self-supervised learning. Kim \etal~\cite{DBLP:journals/corr/abs-2003-08264} and Rakshit \etal~\cite{rakshit2022open} utilized an instance-level self-supervised contrastive learning for representation learning. As the instance-level training is sensitive to outliers, Yue \etal~\cite{DBLP:conf/cvpr/YueZZ0DKS21} utilized the self-supervised learning via class prototypes to enhance model performance and robustness. 
Despite the effectiveness, existing methods generally discard the confidence of different samples, leading to sub-optimal adaptation. Therefore, we propose a novel confidence-based transfer learning method to mitigate such weakness. 

\label{sec:method}

\begin{figure*}[t]
  \centering
   \includegraphics[width=1.0\linewidth]{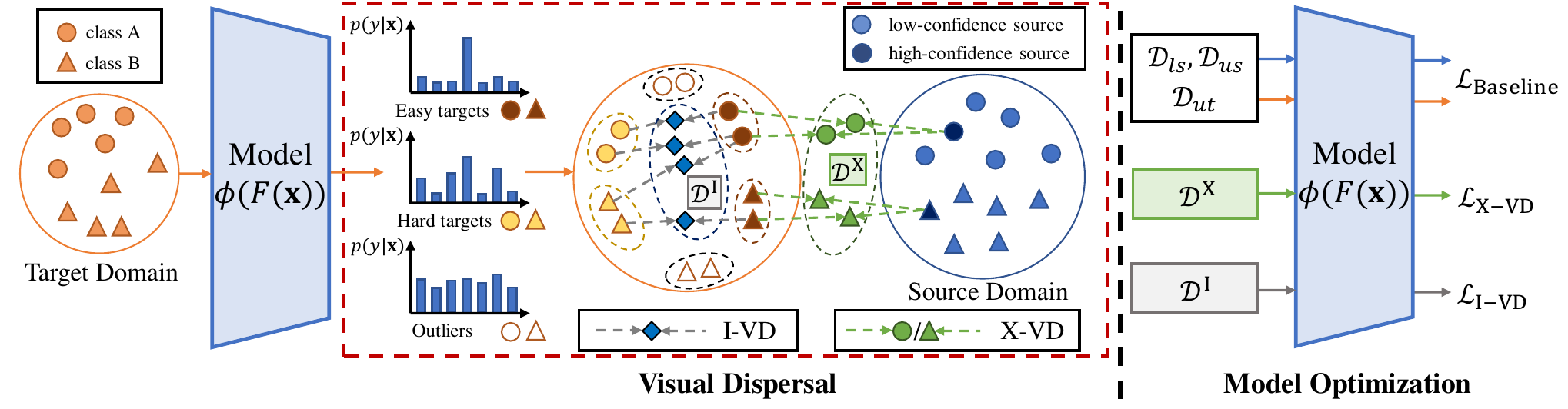}

   \caption{An overview of the proposed C-VisDiT method. X-VD and I-VD strategies are performed across and within domains, respectively, via visual dispersal. We carefully consider the confidence of samples, ending up with a cross-blending dataset $\mathcal{D}^{\text{X}}$ and an intra-mixing dataset $\mathcal{D}^{\text{I}}$. Best viewed in color.}
   \label{fig:model_method}
\end{figure*}

\textbf{MixUp in UDA.}
MixUp~\cite{DBLP:conf/iclr/ZhangCDL18} is an augmentation strategy where linearly interpolated samples are introduced for training. MixUp has been extensively explored in UDA. For example,  
Wu \etal~\cite{wu2020dual} and Xu \etal~\cite{xu2020adversarial} utilize MixUp to create complex samples for adversarial learning. 
Yan \etal~\cite{yan2020improve} and Ding \etal~\cite{ding2022proxymix} directly conduct the MixUp regularization along with the model training process to further boost model performance.
In the FUDA task, to the best of our knowledge,
no previous work has applied MixUp, mainly because the severely sparse annotation (\eg, 1-shot) prohibits providing reliable samples for adaptation, thus harming the capability of MixUp. In our work, we show that with the proposed confidence-based strategy, we can conveniently regain the power of MixUp for FUDA, establishing new state-of-the-art performance on all standard benchmarks for FUDA.


\section{Methodology}

\subsection{Preliminary}
We explore the problem of FUDA in which there are two data domains with different data distributions. As for the source domain, we have sparsely labeled source data $\mathcal{D}_{ls}=\left\{(\mathbf{x}^{ls}_i,y^{ls}_i)\right\}^{N_{ls}}_{i=1}$ and the remaining unlabeled source data $\mathcal{D}_{us}=\left\{(\mathbf{x}^{us}_i)\right\}^{N_{us}}_{i=1}$, where $N_{us}\gg N_{ls}$. For the target domain, only unlabeled data $\mathcal{D}_{ut}=\left\{(\mathbf{x}^{ut}_i)\right\}^{N_{ut}}_{i=1}$ are provided. Following previous works~\cite{DBLP:journals/corr/abs-2003-08264,DBLP:conf/cvpr/YueZZ0DKS21}, we focus on the classification task, where $\mathcal{D}_{ls}$, $\mathcal{D}_{us}$ and $\mathcal{D}_{ut}$ are \textit{all visible} during the model training process. 

\subsection{Baseline Model}
\label{sec:Baseline}
We first introduce a generic solution as our baseline model. Specifically, following~\cite{DBLP:journals/corr/abs-2003-08264,DBLP:conf/cvpr/YueZZ0DKS21}, for a given image $\mathbf{x}$, we leverage a convolutional neural network (CNN), denoted as $F(\cdot)$, to extract its feature. A softmax layer, denoted as $\phi(\cdot)$, is employed as the classifier to predict the semantic distribution $p(y|\mathbf{x})$. The forward process can be formulated as follows:
\begin{equation}
p(y|\mathbf{x}) =  \phi (F(\mathbf{x})) = [p_1,p_2,\cdots,p_c]
\end{equation}
where $c$ is the number of different categories.


During training, for the labeled source data, we employ the standard cross-entropy objective $\mathcal{L}_{CE}$: 
\begin{equation}
\mathcal{L}_{cls} = \sum_j^{N_{ls}} \mathcal{L}_{CE}(p(y_j^{ls}|\mathbf{x}_j^{ls}), y_j^{ls})
\end{equation}



For the unlabeled target data, following~\cite{liang2021source}, we utilize the mutual information maximization objective to encourage individually certain and globally diverse predictions:
\begin{equation}
    \mathcal{L}_{MI} = -\mathcal{H}\left(\sum_{i}^{N_{ut}}(p(y|\mathbf{x}_i^{ut}))\right) + \frac{1}{N_{ut}} \sum_{i}^{N_{ut}} \mathcal{H}(p(y|\mathbf{x}_i^{ut}))
\end{equation}
where the entropy metric $\mathcal{H}(p(y|\mathbf{x}))=\sum_{i=1}^{c}p_i\log p_i$.


In order to enhance the cross-domain feature alignment, we further adopt the prototypical self-supervised learning method~\cite{DBLP:conf/cvpr/YueZZ0DKS21} to optimize the whole network. We denote it as $\mathcal{L}_{self}$ and leave its details to~\cite{DBLP:conf/cvpr/YueZZ0DKS21} due to the page limit.

Finally, the overall learning objective of the baseline model can be derived as:
\begin{equation}
    \mathcal{L}_{\text{Baseline}} = \mathcal{L}_{cls} + \lambda_{MI} \cdot \mathcal{L}_{MI} + \lambda_{self} \cdot \mathcal{L}_{self}
\end{equation}
where $\lambda_{MI}$ and $\lambda_{self}$ are two hyper-parameters.

\subsection{Cross-domain Visual Dispersal (X-VD)}
\label{sec:CDVD}

Transferring knowledge from the source domain to the unlabeled target domain in FUDA is notably difficult, due to the sparsely labeled source data. 
As we observe, in the Clipart-to-Art task with 3\% labeled source on the Office-Home~\cite{venkateswara2017deep} dataset, existing state-of-the-art methods, \ie, CDS~\cite{DBLP:journals/corr/abs-2003-08264} and PCS~\cite{DBLP:conf/cvpr/YueZZ0DKS21}, only achieve 38.7\% and 45.5\% classification accuracy on the {\it unlabeled source dataset}, \ie, $\mathcal{D}_{us}$, respectively. This indicates that 
involving $\mathcal{D}_{us}$ for knowledge transfer inevitably introduces noise to the model adaptation. 
As a result, simply conducting the self-supervised contrastive learning with these unreliable source samples as in~\cite{DBLP:journals/corr/abs-2003-08264,DBLP:conf/cvpr/YueZZ0DKS21} unavoidably leads target samples to be aligned to inappropriate source semantics, thus affecting target classification performance.
Therefore, to remedy such an issue, we propose a cross-domain visual dispersal (\textbf{X-VD}) strategy to enhance the knowledge transfer by concentrating on the high-confidence source samples. 

\textbf{Nearest Source Matching.}
We aim to transfer knowledge from source samples that are highly correlated with the target samples.
For each target sample, we perform a nearest sample matching to associate it with its most similar labeled source sample.
Specifically, we first calculate a score matrix $M=\{m_{ij}\}_{N_{ut}\times N_{ls}}$ in each training epoch. $m_{ij}$ is the confidence score for a pair of samples $(\mathbf{x}^{ut}_i,\mathbf{x}^{ls}_j)$,
which is defined by the Euclidean distance between the corresponding features:
\begin{equation}
    m_{ij} = \Vert F(\mathbf{x}^{ut}_i) - F(\mathbf{x}^{ls}_j) \Vert_2
\end{equation}
Then we derive the nearest source sample for each target sample by finding the column index $\Bar{j}_i$ of the minimum $m_{ij}$ in each row of $M$:
\begin{equation}
    \Bar{j}_i = \arg\min_j (m_{ij})
\end{equation}
With the index $\Bar{j}_i$, we can then locate the nearest labeled source sample for unlabeled target samples, forming a high-confidence source dataset, denoted by $\mathcal{D}_{ut\leftarrow ls}=\left\{(\mathbf{x}^{ut\leftarrow ls}_i,y^{ut\leftarrow ls}_i)\right\}^{N_{ut}}_{i=1}$:
\begin{equation}
    (\mathbf{x}^{ut\leftarrow ls}_i,y^{ut\leftarrow ls}_i) = (\mathbf{x}^{ls}_{\Bar{j}_i},y^{ls}_{\Bar{j}_i})
\end{equation}


\textbf{High-confidence Target Sampling.}
To reduce the impact of mismatched sample pairs which have different semantics, we leave out those unreliable target samples and focus on those with high-confidence.
Specifically, we set $r^{\text{X}}_{h}$ as the fixed ratio of high-confidence target samples and find the corresponding indices of $N_{\text{X}}=r^{\text{X}}_{h}N_{ut}$ target samples with the lowest entropy, formulated as:
\begin{equation}
    O_{conf}=\arg\text{TopK}(-\mathcal{H}(\mathbf{x}^{ut}_i),{N_{\text{X}}})
\end{equation}
where the function $\arg\text{TopK}(\cdot,\cdot)$ returns the indices of $N_{\text{X}}$ largest elements of $-\mathcal{H}(\mathbf{x}^{ut}_i)$.
Then, the high-confidence target set $\mathcal{D}^{conf}_{ut}$ and its corresponding nearest labeled source sample set $\mathcal{D}^{conf}_{ut\leftarrow ls}$ are:
\begin{equation}
    \begin{aligned}
    \mathcal{D}^{conf}_{ut} & =\mathbb{I}(i\in O_{conf})\mathcal{D}_{ut} \\ 
    \mathcal{D}^{conf}_{ut\leftarrow ls} & =\mathbb{I}(i\in O_{conf})\mathcal{D}_{ut\leftarrow ls}
    \end{aligned}
\end{equation}


\textbf{Cross-domain Knowledge Transferring via Visual Dispersal.}
We utilize the MixUp augmentation~\cite{DBLP:conf/iclr/ZhangCDL18} to create a cross-blending dataset $\mathcal{D}^{\text{X}}=\{(\mathbf{x}_i^{\text{X}},y_i^{\text{X}})\}_{i=1}^{N_{\text{X}}}$:
\begin{equation}
\begin{aligned}
    \mathbf{x}_i^{\text{X}}&=\beta \Bar{\mathbf{x}}^{ut}_i + (1-\beta) \Bar{\mathbf{x}}^{ut\leftarrow ls}_i \\ 
    y_i^{\text{X}}&=\beta{y}_{i}^{ut} + (1-\beta)y_{i}^{ut\leftarrow ls}
\end{aligned}
\label{eq:mixup-1-2}
\end{equation}
where ${y}_{i}^{ut}=\arg\max_jp(y|\Bar{\mathbf{x}}^{ut}_i)_j$ is the one-hot pseudo label for $\Bar{\mathbf{x}}^{ut}_i \in \mathcal{D}^{conf}_{ut}$ and $y_{i}^{ut\leftarrow ls}$ is the one-hot ground-truth of $\Bar{\mathbf{x}}^{ut\leftarrow ls}_i \in \mathcal{D}^{conf}_{ut\leftarrow ls}$. $\beta \sim \text{Beta}(\alpha,\alpha)$ is sampled from a beta distribution. As a result, the source knowledge can be well dispersed to the target domain at the visual level, rather than from the latent representation level in CDS~\cite{DBLP:journals/corr/abs-2003-08264} and PCS~\cite{DBLP:conf/cvpr/YueZZ0DKS21}.

Then, our cross-domain visual dispersal loss can be formulated with the cross-entropy loss $\mathcal{L}_{CE}$:
\begin{equation}
    \mathcal{L}_\text{{X-VD}}=\sum_{(\mathbf{x},y)\in \mathcal{D}^{\text{X}}} \mathcal{L}_{CE}(p(y|\mathbf{x}),y)
\end{equation}

\textbf{Discussion.}
Cross-domain visual dispersal transfers only high-confidence source knowledge to target domain, reducing the disturbance of noisy supervisions. During the transfer process, the nearest sample matching strategy intends to match similar samples as pairs. 
Therefore, we can regard each sample in $\mathcal{D}^{\text{X}}$ as a fusion of two independent samples from two similar semantic distributions, although they belong to different domains. Benefited from the visual dispersal, \ie, \cref{eq:mixup-1-2}, the adaptation model can be encouraged to pull two samples with similar semantics but from different domains close to each other, realizing domain-invariant feature learning. As a result, the domain gap can be mitigated with a more reliable knowledge transfer.



\subsection{Intra-domain Visual Dispersal (I-VD)}
\label{sec:IDVD}

\begin{figure}[t]
  \centering
  \includegraphics[width=1.0\linewidth]{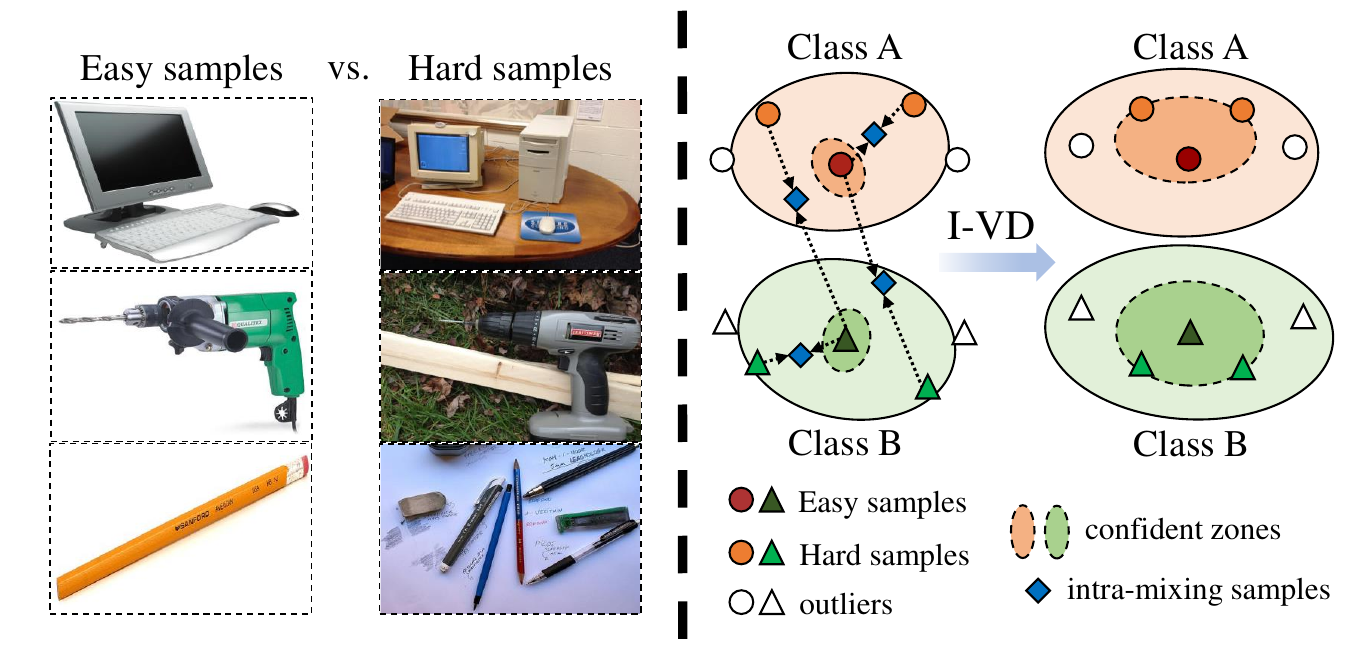}

   \caption{Different target samples present different difficulties to the model learning because of various differences, \eg, object size and background information. Due to the guidance of intra-mixing samples, I-VD gradually fixes the decision boundaries for the target domain, improving the adaptation on hard target samples.}
   \label{fig:decision_bdry}
\end{figure}

Hard samples widely exist during the model learning process~\cite{radenovic2016cnn,DBLP:conf/cvpr/ShrivastavaGG16,xuan2020hard}.
For example, in the Real World domain on the Office-Home dataset~\cite{venkateswara2017deep}, samples with complicated backgrounds and small objects are more likely to have low-confidence probability distributions compared with samples with only one object brimming the image, as shown in \cref{fig:decision_bdry}. Therefore, applying the same learning metrics to all target samples without differentiation makes the model overfit easy target samples, which results in decision boundaries misclassifying hard samples. 
However, existing methods~\cite{DBLP:journals/corr/abs-2003-08264,DBLP:conf/cvpr/YueZZ0DKS21} usually ignore this phenomenon.
To enhance the classification of hard target samples, we propose an intra-domain visual dispersal (\textbf{I-VD}) strategy to guide them with easy ones.

\textbf{Confidence-based Target Domain Splitting.}
The learning difficulty of a target sample is correlated to the confidence of its predicted probability distribution, which can be measured via entropy, as large prediction entropy leads to uncertain model prediction~\cite{wang2016cost}. 
To differentiate these samples, we split the target samples into three groups based on their prediction entropy. The number of samples in each group is denoted as $N_1=r^{\text{I}}_{E}N_{ut}$, $N_2=r^{\text{I}}_{H}N_{ut}$, and $N_3=N_{ut}-N_1-N_2$, respectively, controlled by hyper-parameters $r^{\text{I}}_{E}$ and $r^{\text{I}}_{H}$. Specifically, the indices of samples in three groups are derived as:
\begin{equation}
    \begin{aligned}
        O & = \arg \text{Sort}(\mathcal{H}(p(y|\mathbf{x}^{ut}_i))) \\
        O_1  = O[0\text{:}N_1], \ O_2 & = O[N_1\text{:}N^{'}], \ O_3 = O[N^{'}\text{:}N_{ut}]
    \end{aligned}
\end{equation}
where $\arg \text{Sort}(\cdot)=[i_1,i_2,\cdots,c_n]$ returns the indices that would sort an array in ascending order and $N^{'}=N_1+N_2$. Therefore, we define the easy target sample set $\mathcal{D}_{ut}^{E}$ and the hard target sample set $\mathcal{D}_{ut}^{H}$ as:
\begin{equation}
    \begin{aligned}
        \mathcal{D}_{ut}^{E}=\mathbb{I}(i\in O_1)\mathcal{D}_{ut},\ \ 
        \mathcal{D}_{ut}^{H}=\mathbb{I}(i\in O_2)\mathcal{D}_{ut}
    \end{aligned}
\end{equation}
We exclude the $O_3$ outlier samples $\mathcal{D}_{ut}^{O}$ for training because they are of high uncertainty and thus not trustworthy.

\textbf{Intra-domain Sample Guidance via Visual Dispersal.}
To guide the learning of hard target samples, we disperse the easy target data at the visual level by leveraging the MixUp strategy \cite{DBLP:conf/iclr/ZhangCDL18}. 
We find it helpful to add a small set of high-confidence labeled source samples to the easy target samples 
which enlarges the easy sample set.
For ease of explanation, we denote the final training easy sample set by $\mathcal{T}^E_{ut} = \mathcal{D}_{ls} \cup \mathcal{D}^E_{ut}$, and denote the training hard sample set by $\mathcal{T}^H_{ut}=\mathcal{D}^H_{ut}$.
For each $\Bar{\mathbf{x}}^{H}_i \in \mathcal{T}^H_{ut}$, we choose a random $\Bar{\mathbf{x}}^{E}_j \in \mathcal{T}^E_{ut}$ to guide its learning by conducting image-level MixUp:
\begin{equation}
\begin{aligned}
    \mathbf{x}^{\text{I}}_i&=\beta\Bar{\mathbf{x}}^{E}_j + (1-\beta)\Bar{\mathbf{x}}^{H}_i \\
    y^{\text{I}}_i&=\beta{y}^{E}_j + (1-\beta)y^{H}_i
\end{aligned}
\label{eq:mixup-3-4}
\end{equation}
where one-hot pseudo label $y^{H}_i=\arg\max_k p(y|\Bar{\mathbf{x}}^{H}_i)_k$, and $\beta \sim \text{Beta}(\alpha,\alpha)$ sampled from a beta distribution. For ${y}^{E}_j$, we have:
\begin{align}
    {y}^{E}_j = 
    \begin{cases}
    y^s & \text{if } \Bar{\mathbf{x}}^{E}_j \in \mathcal{D}_{ls}\\
    \arg\max_k p(y|\Bar{\mathbf{x}}^{E}_j)_k & \text{if } \Bar{\mathbf{x}}^{E}_j \in \mathcal{D}_{ut}^{E}
    \end{cases}
\end{align}

With the transformations above, we construct an intra-mixing dataset $\mathcal{D}^{\text{I}}=\{(\mathbf{x}^{\text{I}}_i, y^{\text{I}}_i)\}_{i=1}^{N_{\text{I}}}$, where $N_{\text{I}}=|\mathcal{T}^H_{ut}|$. We directly employ the cross-entropy loss to optimize the intra-mixing samples:
\begin{equation}
    \mathcal{L}_\text{{I-VD}}=\sum_{(\mathbf{x},y)\in \mathcal{D}^{\text{I}}} \mathcal{L}_{CE}(p(y|\mathbf{x}),y)
\end{equation}

\textbf{Discussion.}
The constructed intra-mixing samples act as guidance towards the learning of hard samples.
As illustrated in \cref{fig:decision_bdry}, our intra-mixing samples are located between the easy target samples and the hard ones.
Through learning on the probability distributions of the intra-mixing samples, our model can reduce the undesirable prediction oscillations between the easy samples and the hard ones, as revealed in~\cite{DBLP:conf/iclr/ZhangCDL18}. Due to the fact that model predictions on easy samples are more consistent towards model optimization~\cite{baldock2021deep}, our model in turn enhances the learning of hard samples, thus diminishing the learning disparity between easy samples and hard ones. 
Therefore, our I-VD incisively leads to better performance on hard target samples.

\subsection{C-VisDiT for the FUDA Problem}
Our C-VisDiT learning method mainly introduces cross-domain and intra-domain visual dispersal strategies for the FUDA problem. Together with the baseline training loss, our final training objective is expressed by:
\begin{equation}
    \mathcal{L}=\mathcal{L}_{\text{Baseline}}+\lambda_\text{X-VD}\mathcal{L}_\text{{X-VD}}+\lambda_\text{I-VD}\mathcal{L}_\text{{I-VD}}
\end{equation}

\begin{table*}[t]
\centering
\caption{Adaptation accuracy (\%) comparison on 1-shot / 3-shots labeled source per class on the Office-31 dataset.}
\resizebox{0.8\textwidth}{!}{
\begin{threeparttable}
\begin{tabular}{l|c|c|c|c|c|c|c}
\toprule
 Method & A$\rightarrow$D & A$\rightarrow$W & D$\rightarrow$A & D$\rightarrow$W & W$\rightarrow$A & W$\rightarrow$D & Avg \\
\midrule
Source Only & 31.3 / 49.0 & 19.6 / 43.3 & 41.3 / 55.7 & 58.7 / 81.8 & 39.9 / 49.8 & 61.9 / 81.7 & 42.1 / 60.2 \\
MME~\cite{saito2019semi} & 21.5 / 51.0 & 12.2 / 54.6 & 23.1 / 60.2 & 60.9 / 89.7 & 14.0 / 52.3 & 62.4 / 91.4 & 32.3 / 66.5 \\
CDAN~\cite{long2018conditional}   & 11.2 / 43.7 & 6.2 / 50.1  & 9.1 / 65.1  & 54.8 / 91.6 & 10.4 / 57.0 & 41.6 / 89.8 & 22.2 / 66.2 \\
MDDIA~\cite{jiang2020implicit} & 45.0 / 62.9 & 54.5 / 65.4 & 55.6 / 67.9 & 84.4 / 93.3 & 53.4 / 70.3 & 79.5 / 93.2 & 62.1 / 75.5 \\
CDS~\cite{DBLP:journals/corr/abs-2003-08264} &  52.6 / 65.1  &  55.2 / 68.8  &  65.7 / 71.2  &  76.6 / 88.1  &  59.7 / 71.0  &  73.3 / 87.3  &  63.9 / 75.3 \\
PCS~\cite{DBLP:conf/cvpr/YueZZ0DKS21} & {60.2} / {78.2} & {69.8} / {82.9} & \textbf{76.1} / {76.4} & {90.6} / {94.1} & {71.2} / {76.3} & {91.8} / {96.0} & {76.6} / {84.0} \\
\midrule
C-VisDiT (Ours) & \textbf{74.1} / \textbf{82.7} & \textbf{72.3} / \textbf{86.0} & {75.7} / \textbf{76.5} & \textbf{93.2} / \textbf{95.0} & \textbf{76.4} / \textbf{76.9} & \textbf{94.2} / \textbf{97.0} & \textbf{81.0} / \textbf{85.7} \\
\bottomrule
\end{tabular}
\end{threeparttable}
}
\label{tab:office}
\end{table*}

\begin{table*}[]
\centering
\caption{Adaptation accuracy (\%) comparison on 3\% and 6\% labeled source samples per class on the Office-Home dataset.}
\resizebox{0.97\textwidth}{!}{%
\begin{threeparttable}
\begin{tabular}{l|c|c|c|c|c|c|c|c|c|c|c|c|c}
\toprule[1.5pt]
 Method & Ar $\rightarrow$Cl & Ar $\rightarrow$Pr & Ar $\rightarrow$Rw & Cl $\rightarrow$Ar & Cl $\rightarrow$Pr & Cl $\rightarrow$Rw & Pr $\rightarrow$Ar & Pr $\rightarrow$Cl & Pr $\rightarrow$Rw & Rw $\rightarrow$Ar & Rw $\rightarrow$Cl & Rw $\rightarrow$Pr & Avg \\
 \midrule
 \multicolumn{14}{c}{\textbf{3\% labeled source}} \\ \midrule
Source Only & 22.5 & 36.5 & 41.1 & 18.5 & 29.7 & 28.6 & 27.2 & 25.9 & 38.4 & 33.5 & 20.3 & 41.4 & 30.3 \\
MME~\cite{saito2019semi} & 4.5 & 15.4 & 25.0 & 28.7 & 34.1 & 37.0 & 25.6 & 25.4 & 44.9 & 39.3 & 29.0 & 52.0 & 30.1 \\
CDAN~\cite{long2018conditional} & 5.0 & 8.4 & 11.8 & 20.6 & 26.1 & 27.5 & 26.6 & 27.0 & 40.3 & 38.7 & 25.5 & 44.9 & 25.2 \\
MDDIA~\cite{jiang2020implicit} & 21.7 & 37.3 & 42.8 & 29.4 & 43.9 & 44.2 & 37.7 & 29.5 & 51.0 & 47.1 & 29.2 & 56.4 & 39.1 \\
CDS~\cite{DBLP:journals/corr/abs-2003-08264} & {43.8} & 55.5 & 60.2 & 51.5 & 56.4 & 59.6 & 51.3 & 46.4 & 64.5 & 62.2 & {52.4} & 70.2 & 56.2 \\
PCS~\cite{DBLP:conf/cvpr/YueZZ0DKS21} & 42.1 & {61.5} & {63.9} & {52.3} & {61.5} & {61.4} & {58.0} & {47.6} & {73.9} & {66.0} & \textbf{52.5} & {75.6} & {59.7} \\ 
\hline
C-VisDiT (Ours) & \textbf{44.1} & \textbf{66.8} & \textbf{67.0} & \textbf{54.9} & \textbf{66.4} & \textbf{66.8} & \textbf{60.5} & \textbf{47.9} & \textbf{75.7} & \textbf{67.2} & 51.6 & \textbf{78.8} & \textbf{62.3} \\ 
\hline \midrule

 \multicolumn{14}{c}{\textbf{6\% labeled source}} \\ \midrule
Source Only & 26.5 & 41.3 & 46.7 & 29.3 & 40.4 & 37.9 & 35.5 & 31.6 & 57.2 & 46.2 & 32.7 & 59.2 & 40.4 \\
MME~\cite{saito2019semi} & 27.6 & 43.2 & 49.5 & 41.1 & 46.6 & 49.5 & 43.7 & 30.5 & 61.3 & 54.9 & 37.3 & 66.8 & 46.0 \\
CDAN~\cite{long2018conditional} & 26.2 & 33.7 & 44.5 & 34.8 & 42.9 & 44.7 & 42.9 & 36.0 & 59.3 & 54.9 & 40.1 & 63.6 & 43.6 \\
MDDIA~\cite{jiang2020implicit} & 25.1 & 44.5 & 51.9 & 35.6 & 46.7 & 50.3 & 48.3 & 37.1 & 64.5 & 58.2 & 36.9 & 68.4 & 50.3 \\
CDS~\cite{DBLP:journals/corr/abs-2003-08264} & 45.4 & 60.4 & 65.5 & 54.9 & 59.2 & 63.8 & 55.4 & \textbf{49.0} & 71.6 & 66.6 & \textbf{54.1} & 75.4 & 60.1 \\
PCS~\cite{DBLP:conf/cvpr/YueZZ0DKS21} & {46.1} & {65.7} & {69.2} & {57.1} & {64.7} & {66.2} & {61.4} & 47.9 & {75.2} & {67.0} & {53.9} & {76.6} & {62.6} \\
\hline
C-VisDiT (Ours) & \textbf{46.5} & \textbf{69.8} & \textbf{72.5} & \textbf{60.2} & \textbf{71.2} & \textbf{71.2} & \textbf{64.1} & \textbf{49.0} & \textbf{78.5} & \textbf{69.1} & 52.8 & \textbf{80.1} & \textbf{65.4} \\
\bottomrule
\end{tabular}%
\end{threeparttable}
}
\label{tab:officehome}
\vspace{-2mm}
\end{table*}

\begin{table}[]
\centering
\caption{Adaptation accuracy (\%) comparison on 1\% labeled source samples per class on the VisDA-C dataset. ``SO'' denotes training with only labeled source samples (Source Only).}
\vspace{-1mm}
\resizebox{1.0\columnwidth}{!}{%
\begin{threeparttable}
\setlength{\tabcolsep}{1.0mm}{
\begin{tabular}{l|ccccccc}
\toprule[1.0pt]
  Method & SO & MME~\cite{saito2019semi} & CDAN~\cite{long2018conditional} & CDS~\cite{DBLP:journals/corr/abs-2003-08264} & PCS~\cite{DBLP:conf/cvpr/YueZZ0DKS21} &  C-VisDiT \\
\midrule
 Acc. & 42.6 & 69.4 & 61.5 & 69.4 & 79.0 & \textbf{80.5} \\
\bottomrule
\end{tabular}}%
\end{threeparttable}
}
\label{tab:visda}
\vspace{-2mm}
\end{table}

\begin{table}[]
\centering
\caption{Adaptation accuracy (\%) comparison on 1-shot and 3-shots labeled source per class on the DomainNet dataset. ``SO'' denotes training with only labeled source samples (Source Only).}
\vspace{-1mm}
\resizebox{0.87\columnwidth}{!}{%
\begin{threeparttable}
\setlength{\tabcolsep}{1.0mm}{
\begin{tabular}{l|cccc}
\toprule[1.0pt]
  Method & SO & MME~\cite{saito2019semi} & CDAN~\cite{long2018conditional} & CDS~\cite{DBLP:journals/corr/abs-2003-08264} \\
\midrule
 1-shot & 13.1 & 17.5 & 14.6 & 21.5 \\ 
 3-shots & 27.1 & 29.1 & 27.3 & 42.5 \\ \midrule
 Method & PCS~\cite{DBLP:conf/cvpr/YueZZ0DKS21} & C-VisDiT & BrAD~\cite{harary2022unsupervised} & BrAD+C-VisDiT \\ \midrule
 1-shot & 34.7 & \textbf{35.9} & 49.7 & \textbf{51.0} \\
 3-shots & 46.1 & \textbf{48.1} & 60.9 & \textbf{63.4} \\
\bottomrule
\end{tabular}}%
\end{threeparttable}
}
\label{tab:domainnet}
\vspace{-2mm}
\end{table}

\section{Experiments}

\subsection{Experimental Setting}
\label{sec:set}

\textbf{Datasets.}
We evaluate our method on four widely-used benchmark datasets for FUDA, \ie, {\bf Office-31}~\cite{saenko2010adapting}, {\bf Office-Home}~\cite{venkateswara2017deep}, {\bf VisDA-C}~\cite{peng2017visda}, and {\bf DomainNet}~\cite{peng2019moment}, following prior works~\cite{DBLP:journals/corr/abs-2003-08264,DBLP:conf/cvpr/YueZZ0DKS21}. For Office-31, We conduct experiments with 1-shot
and 3-shots source labels per class. For Office-home, we employ 3\% and 6\% labeled source images per class. For VisDA-C, we validate our model
with 1\% labeled source images per class. For DomainNet, we employ 1-shot and 3-shots source labels per class. Statistics of benchmarks can be found in the supplementary materials.

For fair comparison, we implement our model following PCS~\cite{DBLP:conf/cvpr/YueZZ0DKS21} and BrAD~\cite{harary2022unsupervised}. More implementation details are provided in the supplementary materials.

\subsection{Comparison with State-of-the-art Methods}
\label{sec:res_sota}


We choose the baseline methods as PCS~\cite{DBLP:conf/cvpr/YueZZ0DKS21}, including MME~\cite{saito2019semi}, CDAN~\cite{long2018conditional}, MDDIA~\cite{jiang2020implicit}, and CDS~\cite{DBLP:journals/corr/abs-2003-08264}.
Extensive experiments are conducted on the Office-31, Office-Home, VisDA-C, and DomainNet datasets, with the corresponding results in terms of classification accuracy reported in \cref{tab:office}, \cref{tab:officehome}, \cref{tab:visda}, and \cref{tab:domainnet}, respectively. On the DomainNet dataset, we also initialize the model backbone with generalization weights pre-trained on four unlabeled domains provided in BrAD~\cite{harary2022unsupervised}, and then train the model with our C-VisDiT method, denoted as BrAD+C-VisDiT. 
We can observe that the proposed C-VisDiT can significantly outperform existing methods on all benchmarks. 
Specifically, compared to the existing state-of-the-art method PCS, C-VisDiT improves the classification accuracy on the Office-31 dataset by 4.4\% (1-shot) and 1.7\% (3-shots). 
On the Office-Home dataset, C-VisDiT achieves accuracy gains of 2.6\% (3\% labeled source) and 2.8\% (6\% labeled source). 
On the VisDA-C dataset, C-VisDiT achieves a performance improvement of 1.5\%. 
On the DomainNet dataset, C-VisDiT can achieve 1.2\%/2.0\% and 1.3\%/2.5\% accuracy gain (1-shot/3-shots) compared to PCS and BrAD, respectively. 
Looking into each adaptation setting inside both datasets, 
our method can accomplish a maximum performance improvement of 13.9\% (1-shot setting in A$\rightarrow$D on the Office-31 dataset).
These results show that the proposed C-VisDiT establishes new state-of-the-art performance for FUDA, well demonstrating its effectiveness. More results and details can be found in the supplementary materials.

\subsection{Model Analysis}
\label{sec:ablation}

In this section, we present the ablation studies of our proposed C-VisDiT. We also give more investigations of our X-VD and I-VD strategies. For the convenience of analysis, we construct two C-VisDiT variants, \ie, C-VisDiT-X and C-VisDiT-I, trained with X-VD and I-VD only, respectively. 
More analysis, \eg the hyper-parameter analysis, can be found in the supplementary materials.
Following PCS~\cite{DBLP:conf/cvpr/YueZZ0DKS21}, here we employ the Office-31 dataset to inspect our model.

\textbf{Ablation Studies.}
Here we investigate the effect of the proposed X-VD and I-VD strategies in C-VisDiT. 
The results are reported in \cref{tab:office_ablation_main}.
We can see that all three C-VisDiT variants can consistently outperform the baseline model by a great margin. Meanwhile, compared with the previous state-of-the-art method, \ie, PCS~\cite{DBLP:conf/cvpr/YueZZ0DKS21}, both C-VisDiT-X and C-VisDiT-I can achieve better results. As a combination of X-VD and I-VD, C-VisDiT significantly exceeds the PCS with a maximum performance improvement of 4.4\% (1-shot setting on the Office-31).
These results show that our X-VD and I-VD strategies are effective and complementary.


\begin{table}[]
\centering
\caption{Performance contribution of each component on the Office-31 dataset in terms of adaptation accuracy (\%).}
\resizebox{0.7\columnwidth}{!}{%
\begin{tabular}{c|cc|c}
\toprule
{Method} & {$\mathcal{L}_{\text{X-VD}}$} & {$\mathcal{L}_{\text{I-VD}}$} & 1-shot / 3-shots  \\ 
\midrule
PCS~\cite{DBLP:conf/cvpr/YueZZ0DKS21} & - & - & 76.6 / 84.0 \\
\midrule
Baseline & $\times$ & $\times$ & 77.6 / 83.8 \\
C-VisDiT-X & $\surd$ & $\times$ & 79.2 / 84.9 \\
C-VisDiT-I & $\times$ & $\surd$ & 78.8 / 85.0 \\
C-VisDiT & $\surd$ & $\surd$ & \textbf{81.0} / \textbf{85.7} \\
\bottomrule
\end{tabular}%
}
\label{tab:office_ablation_main}
\end{table}


\begin{table}[]
\centering
\caption{Adaptation accuracy for C-VisDiT-X with different confidence-based strategies on the Office-31 dataset (\%).}
\resizebox{0.75\columnwidth}{!}{%
\begin{tabular}{c|c|c}
\toprule
 Method & Unlabeled source & 1-shot / 3-shots \\
\midrule
 Baseline & - & 77.6 / 83.8 \\
 \midrule
 \multirow{2}{*}{MixUp} & 100\% & 74.8 / 82.6 \\
 & 0\% \textbf{(Ours)} & 78.6 / 84.5 \\
 \midrule
 \multirow{4}{*}{C-VisDiT-X} & 100\% & 74.8 / 82.4 \\
 & 10\% & 76.1 / 83.8 \\
 & 1\% & 76.3 / 84.2 \\
 & 0\% \textbf{(Ours)} & \textbf{79.2} / \textbf{84.9} \\
\bottomrule
\end{tabular}%
}
\label{tab:office_implement_CDVD}
\end{table}

\textbf{Superiority of Confidence-based Knowledge Transfer in X-VD.}
The confidence-based strategies play a vital role in the proposed X-VD. 
First, we demonstrate the effectiveness of our confidence-based strategies by validating on the C-VisDiT-X variant.  
We directly conduct MixUp as a compared baseline, in which for a given target sample, a random source sample is selected for MixUp. 
As shown in \cref{tab:office_implement_CDVD}, MixUp with 100\% unlabeled source samples performs significantly worse than the baseline. This shows that plainly conducting MixUp severely harms the performance. However, when using only high-confidence knowledge with 0\% unlabeled source, MixUp can easily outperform both the baseline and the 100\% unlabeled source variant. This shows that MixUp is greatly restrained due to the massive noise in FUDA setting and eliminating the noise is crucial in the proposed X-VD. 
Furthermore, replacing MixUp (0\% unlabeled source) with our C-VisDiT-X achieves the best result, well demonstrating the effectiveness of our confidence-based strategies. 
Therefore, the superiority of our X-VD can largely attribute to the confidence-based knowledge transfer, rather than the MixUp transformations.

Second, we investigate the impact of source knowledge purity. We compare C-VisDiT-X performance 
by further involving 100\%, 10\%, and 1\% unlabeled source samples. As shown in \cref{tab:office_implement_CDVD}, limiting the source samples to only labeled source samples (0\% unlabeled source) gives the best performance, while including 100\%, 10\%, and 1\% unlabeled source samples all lead to performance drops. This indicates that the high purity of source knowledge is very helpful for the cross-domain knowledge transfer. 

\begin{table}[]
\centering
\caption{Adaptation accuracy of different target sample guidance strategies for C-VisDiT-I on the Office-31 dataset (\%).}
\resizebox{1.0\columnwidth}{!}{%
\begin{tabular}{c|cc|c}
\toprule
 Method & $\mathcal{T}^{E}_{ut}$ & $\mathcal{T}^{H}_{ut}$ & 1-shot / 3-shots \\
\midrule
 Baseline & - & - & 77.6 / 83.8 \\
 \midrule
 MixUp & $\mathcal{D}_{ut}$ & $\mathcal{D}_{ut}$ & 77.3 / 83.9 \\
 \midrule
 \multirow{4}{*}{C-VisDiT-I} & $\mathcal{D}_{ut}^{E}$ & $\mathcal{D}_{ut}^{H}\cup \mathcal{D}_{ut}^{O}$ & 77.7 / 84.5 \\
 & $\mathcal{D}_{ut}^{E}$ & $\mathcal{D}_{ut}^{H}$ \textbf{(Ours)} & 78.2 / 84.4 \\
 & $\mathcal{D}_{ls}\cup \mathcal{D}_{ut}^{E}$ \textbf{(Ours)} & $\mathcal{D}_{ut}^{H}\cup \mathcal{D}_{ut}^{O}$ & 78.3 / 84.6 \\
 & $\mathcal{D}_{ls}\cup \mathcal{D}_{ut}^{E}$ \textbf{(Ours)} & $\mathcal{D}_{ut}^{H}$ \textbf{(Ours)} & \textbf{78.8} / \textbf{85.0} \\
\bottomrule
\end{tabular}%
}
\label{tab:office_implement_IDVD}
\end{table}

\textbf{Necessity of Confidence-based Target Sample Guidance in I-VD.}
To prove the necessity and superiority of our I-VD strategy for model adaptation, we compare a series of sample guidance strategies.
First, we compare our proposed I-VD with applying random MixUp on the entire target domain. 
As shown in \cref{tab:office_implement_IDVD}, plainly conducting MixUp barely changes target performance, while all our I-VD variants boosted performance. This shows that, although our I-VD is based on MixUp, our confidence-based target sample guidance greatly improves target performance. 

Furthermore, we dive into the designs of our confidence-based strategies, and investigate how the choice of $\mathcal{T}^{E}_{ut}$ and $\mathcal{T}^{H}_{ut}$ affects target performance. Results show that both adding confident labeled source samples $\mathcal{D}_{ls}$ to $\mathcal{T}^{E}_{ut}$ and excluding low-confidence target outliers $\mathcal{D}^{O}_{ut}$ for $\mathcal{T}^{H}_{ut}$ lead to performance gains, and combining the strategies would boost the performance even more, further demonstrating the effectiveness of confidence target sample guidance. 

\begin{figure}[t]
\centering
    \includegraphics[width=0.89\linewidth, trim={0 0cm 0cm 1.0cm}, clip]{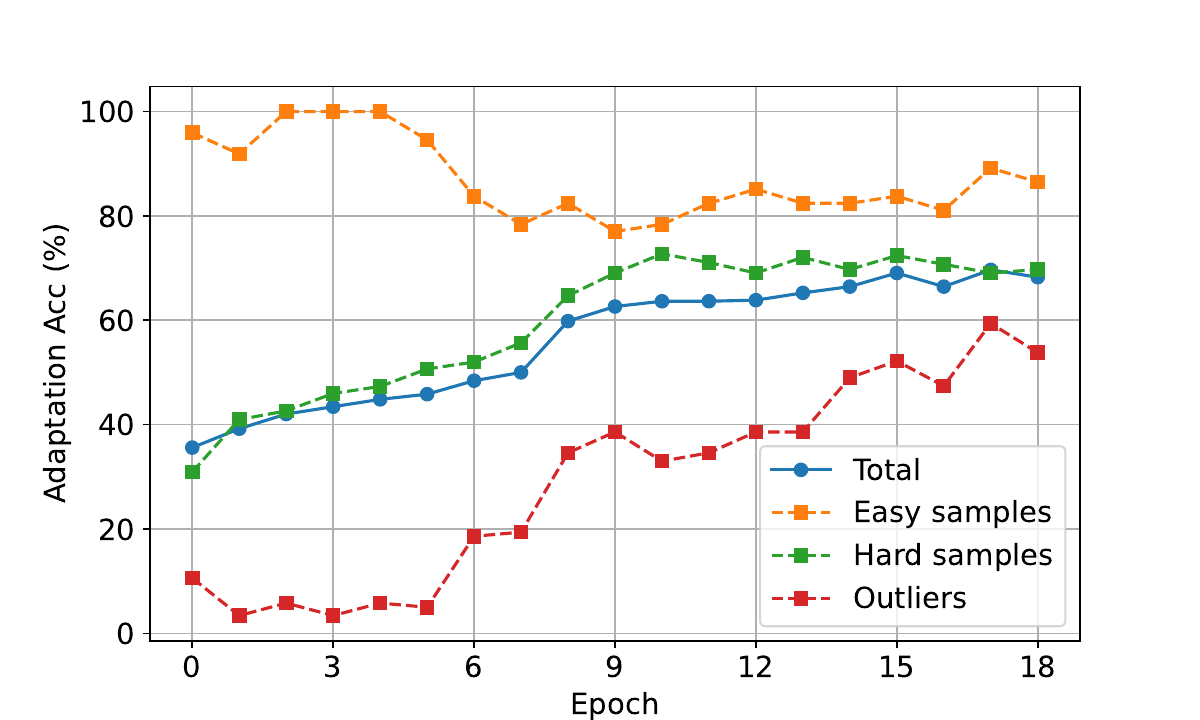}
    \caption{Accuracy change of target samples with different learning difficulties in A$\rightarrow$D (1-shot) on the Office-31 dataset.}
\label{fig:confidence}
\end{figure}

Besides, we qualitatively show the guidance feature of our I-VD strategy towards target samples with different learning difficulties. As illustrated in \cref{fig:confidence}, I-VD improves adaptation accuracy on hard target samples and outliers dramatically. The slight accuracy loss on easy target samples is due to the model overfitting these samples at the beginning. This proves that I-VD is capable of enhancing the model learning on target samples with consideration of their different learning difficulties. 

\begin{table}[]
\centering
\caption{Effect of the confidence-based strategies in C-VisDiT on the Office-31 dataset in terms of adaptation accuracy (\%).}
\resizebox{0.58\columnwidth}{!}{%
\begin{tabular}{c|c}
\toprule
Implementation & 1-shot / 3-shots \\
\midrule
Baseline &  77.6 / 83.8 \\
Plain MixUp &  75.1 / 83.0 \\
C-VisDiT & \textbf{81.0} / \textbf{85.7}  \\
\bottomrule
\end{tabular}%
}
\label{tab:office_confidence}
\end{table}

\textbf{Superiority of C-VisDiT over MixUp.} 
Motivated by the fact that samples with low prediction confidence commonly exist in FUDA, we propose the confidence-based strategies to comprehensively consider sample confidence during training.
To prove the effectiveness of confidence-based strategies more directly, we introduce plain MixUp based on C-VisDiT, in which X-VD is substituted by MixUp on a given target sample and a random source sample, and I-VD is substituted by random MixUp on all target samples. 
As shown in \cref{tab:office_confidence}, plain MixUp underperforms the baseline model, while our C-VisDiT greatly outperforms the baseline model, indicating that the proposed confidence-based strategies are the dominant factor to the superior performance.
Therefore, our C-VisDiT is more suitable for FUDA compared to vanilla MixUp. 

\begin{figure}[htbp]
  \centering
\begin{subfigure}{.5\columnwidth}
    \centering
    \includegraphics[width=0.85\linewidth, trim={0 0 0 0}, clip]{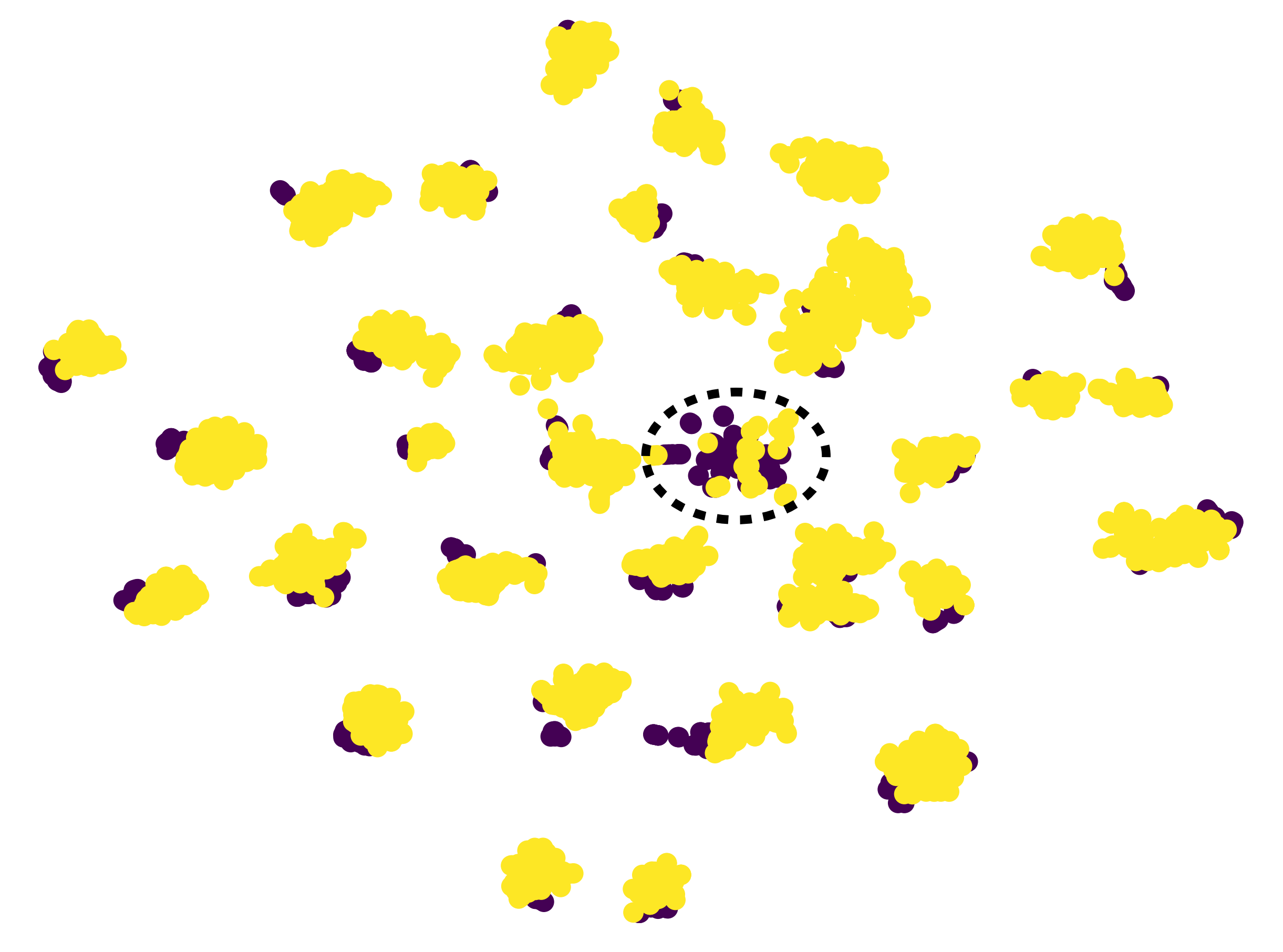}
    \centering
    \caption*{PCS}
\end{subfigure}%
\begin{subfigure}{.5\columnwidth}
    \centering
    \includegraphics[width=0.85\linewidth, trim={0 0 0 0}, clip]{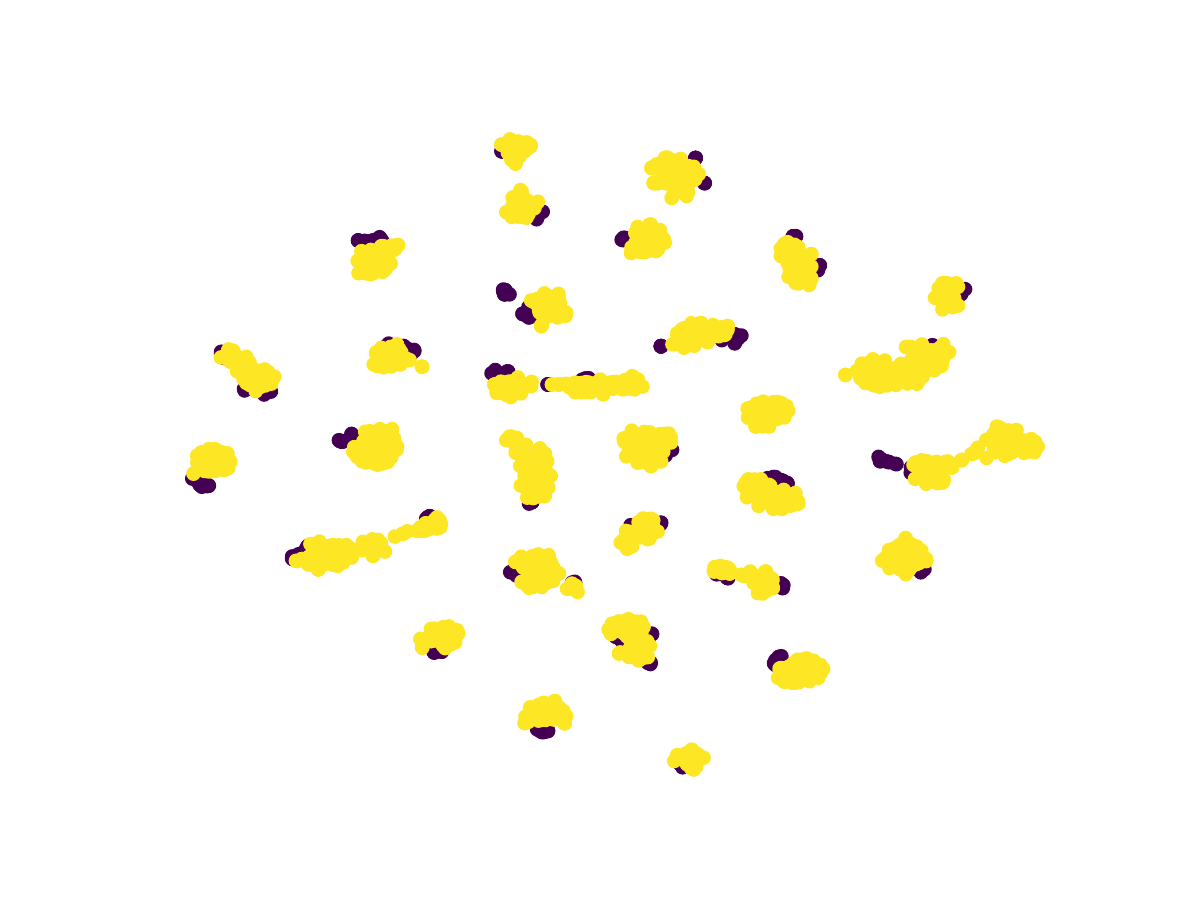}
    \centering
    \caption*{C-VisDiT-X}
\end{subfigure}
\begin{subfigure}{.5\columnwidth}
    \centering
    \includegraphics[width=0.85\linewidth, trim={0 0 0 0}, clip]{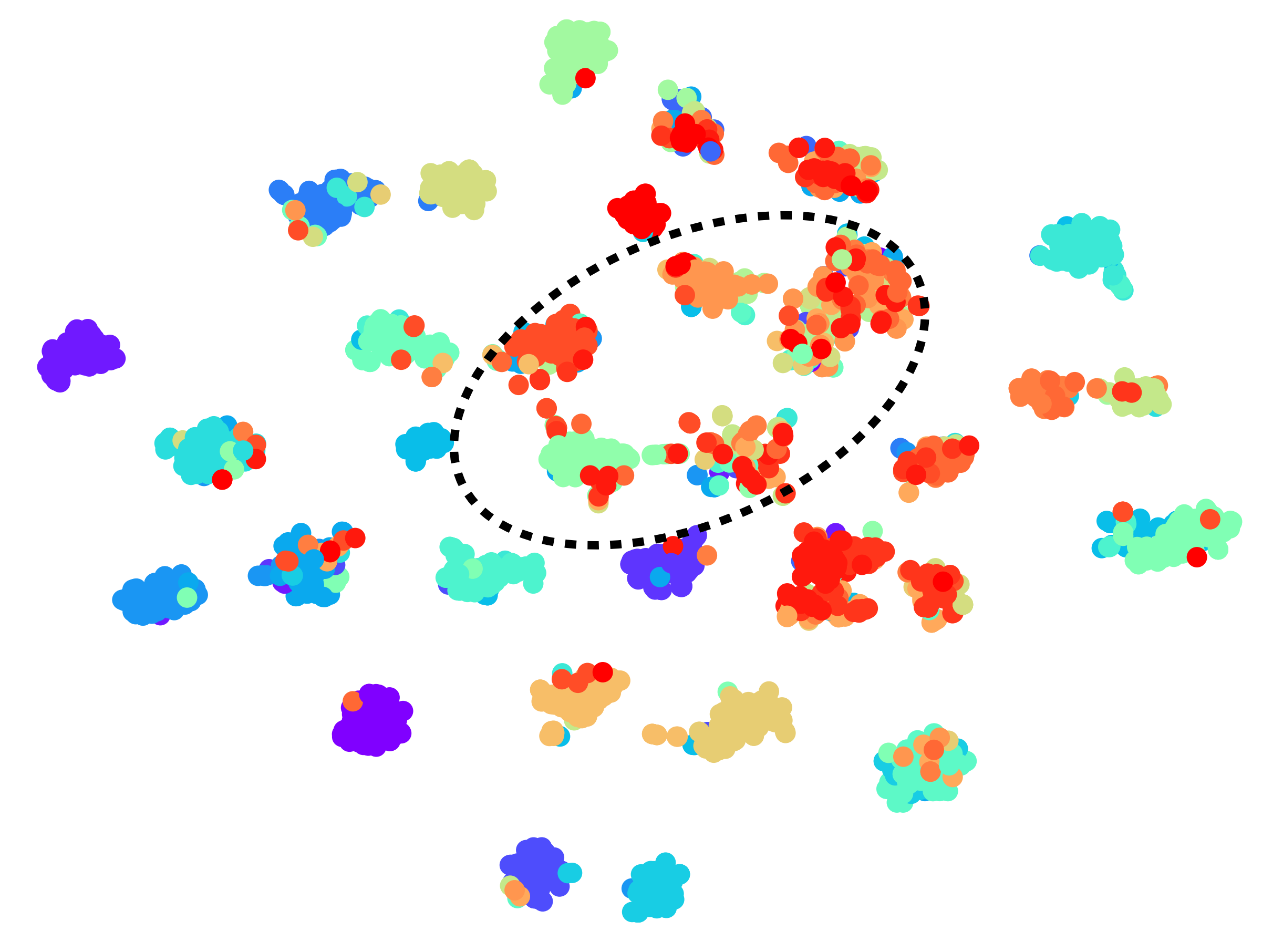}
    \centering
    \caption*{PCS}
\end{subfigure}%
\begin{subfigure}{.5\columnwidth}
    \centering
    \includegraphics[width=0.85\linewidth, trim={0 0 0 0}, clip]{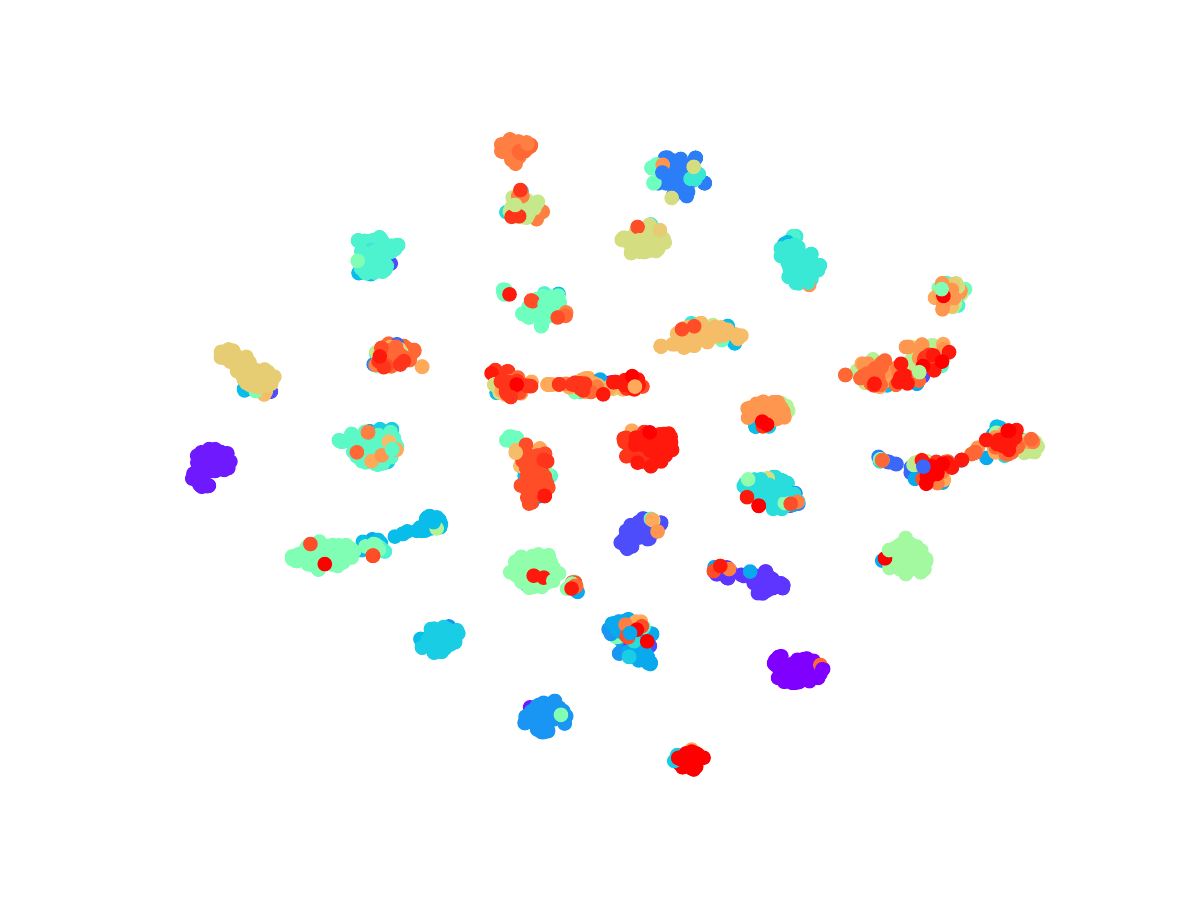}
    \centering
    \caption*{C-VisDiT-I}
\end{subfigure} \\ %
   \caption{t-SNE~\cite{van2008visualizing} visualization results of C-VisDiT-X and C-VisDiT-I compared to PCS~\cite{DBLP:conf/cvpr/YueZZ0DKS21} in A$\rightarrow$D (1-shot) setting on the Office-31 dataset. Top row: \textcolor{tsne_yellow}{\textbf{Yellow}} represents target features and \textcolor{tsne_purple}{\textbf{Purple}} represents source features. Bottom row: different colors represent different sample classes. Best viewed in color.}
   \label{fig:tSNE}
\end{figure}

\textbf{Visualization results.}
To qualitatively analyze the effect of the proposed X-VD and I-VD, we visualize the image features extracted by C-VisDiT-X and C-VisDiT-I via t-SNE~\cite{van2008visualizing}. As shown in \cref{fig:tSNE},
compared to PCS~\cite{DBLP:conf/cvpr/YueZZ0DKS21}, C-VisDiT-X matches the target samples to the source samples better and C-VisDiT-I produces better clusters.
These observations support that our proposed method can construct a better feature space for the model adaptation.

\section{Conclusion}
In this paper, we cope with Few-shot Unsupervised Domain Adaptation (FUDA) in which only a small set of source domain samples are labeled, while all samples in the target domain are unlabeled. 
We propose a confidence-based novel visual dispersal transfer learning method (C-VisDiT) that transfers knowledge only from high-confidence samples to low-confidence samples via 
cross-domain visual dispersal (X-VD) and intra-domain visual dispersal (I-VD).
The proposed X-VD enhances cross-domain knowledge transfer and mitigates the domain gap, while the proposed I-VD guides the learning of hard target samples and boosts the model adaptation.   
Extensive experiments conducted on various domain adaptation benchmarks show that our C-VisDiT outperforms existing methods, establishing new state-of-the-art performance for FUDA. 
In the future, we plan to extend our method to semantic segmentation, object detection, etc.

\section{Acknowledgements}
This research was supported by National Key R\&D Program of China (2021ZD0114703), National Natural Science Foundation of China (Nos. 61925107, U1936202, 62271281, 62021002), Zhejiang Provincial Natural Science Foundation of China (No. LDT23F01013F01), and CCF-DiDi GAIA Collaborative Research Funds for Young Scholars.

{\small
\bibliographystyle{ieee_fullname}
\bibliography{egbib}
}

\clearpage
\appendix
\noindent\textbf{\Large Appendix}

\input{supp_final}

\end{document}

%% file: supp_final.tex
\section{Additional Datasets Details}

We present the overall statistics of the datasets that are used to evaluate our proposed C-VisDiT in \cref{tab:dataset}. {\bf Office-31}~\cite{saenko2010adapting} is a dataset of office images containing 3 domains (Amazon, DSLR, and Webcam) with 31 classes. {\bf Office-Home}~\cite{venkateswara2017deep} also contains office images across 4 domains (Art, Clipart, Product, Real) with 65 classes in each domain. {\bf VisDA-C}~\cite{peng2017visda} is a large simulation-to-real dataset with over 150K images in the training domain and over 55K images in the validation domain. {\bf DomainNet}~\cite{peng2019moment} is the most diverse and recent cross-domain benchmark to-date. Following~\cite{DBLP:conf/cvpr/YueZZ0DKS21,harary2022unsupervised}, we use a subset of DomainNet containing 4 domains (Clipart, Real, Painting and Sketch) and 126 classes. We follow the same settings in~\cite{DBLP:journals/corr/abs-2003-08264,DBLP:conf/cvpr/YueZZ0DKS21,harary2022unsupervised} and conduct experiments with corresponding numbers of labeled images on the aforementioned datasets. For labeled image sampling, we directly leverage the split files\footnote{https://github.com/zhengzangw/PCS-FUDA} published by ~\cite{DBLP:conf/cvpr/YueZZ0DKS21} for a fair comparison with the existing FUDA methods. 


\begin{table}[t]
\centering
\caption{Overall statistics of datasets used to evaluate the proposed C-VisDiT method.}
\label{tab:dataset}
\resizebox{1.0\columnwidth}{!}{
\begin{tabular}{cccccc}
\toprule[2pt]
Dataset & Domain &  Total images & \multicolumn{2}{c}{Labeled images}  &  Classes \\ \midrule[1.5pt]
 & & & 1-shot  & 3-shots & \\\cline{1-6}
\multirow{3}{*}{Office-31~\cite{saenko2010adapting}} & Amazon (A) & 2817 & 31 & 93 & \multirow{3}{*}{31} \\ \cline{2-5}
 & DSLR (D) & 498 & 31 & 93 & \\ \cline{2-5}
 & Webcam (W) & 795 & 31 & 93 & \\ \midrule[1.5pt]
 & & & 3\% & 6\% & \\\cline{1-6}
\multirow{4}{*}{Office-Home~\cite{venkateswara2017deep}} & Art (Ar) & 2427 & 73 & 146 & \multirow{4}{*}{65} \\ \cline{2-5}
 & Clipart (Cl) & 4365 & 131 & 262 & \\ \cline{2-5}
 & Product (Pr) & 4439 & 133 & 266 & \\ \cline{2-5}
 & Real (Rw) & 4357 & 131 & 261 & \\ \midrule[1.5pt]
 & & & \multicolumn{2}{c}{1\%} & \\ \cline{1-6}
\multirow{2}{*}{VisDA-C~\cite{peng2017visda}} & Train & 152397 & \multicolumn{2}{c}{1531} & \multirow{2}{*}{12} \\ \cline{2-5}
 & Validation & 55388 & \multicolumn{2}{c}{-} &  \\ \midrule[1.5pt]
 & & & {1-shot} & 3-shots & \\ \cline{1-6}
\multirow{4}{*}{DomainNet~\cite{peng2019moment}} & \multicolumn{1}{c}{Clipart (C)} & 18703 & {126} & 378 & \multirow{4}{*}{126} \\ \cline{2-5}
 & \multicolumn{1}{c}{Painting (P)} & 31502 & {126} & 378 &  \\ \cline{2-5}
 & \multicolumn{1}{c}{Real (R)} & 70358 & {126} & 378 &  \\ \cline{2-5}
 & \multicolumn{1}{c}{Sketch (S)} & 24582 & {126} & 378 &  \\ \bottomrule[1.5pt]
\end{tabular}
}
\end{table}


\section{Additional Implementation Details}

\subsection{Details for the Baseline Model}

To enhance the cross-domain feature alignment, we adopt the prototypical self-supervised learning method in~\cite{DBLP:conf/cvpr/YueZZ0DKS21} to optimize the baseline model and denote the objective as $\mathcal{L}_{self}$ in the article. Following~\cite{DBLP:conf/cvpr/YueZZ0DKS21}, we utilize the k-means clustering to obtain normalized source class centers $\{\mathbf{\mu}^s_{i}\}_{i=1}^{c}$ and normalized target class centers $\{\mathbf{\mu}^t_{i}\}_{i=1}^{c}$. The class centers are updated via momentum in each training epoch. For each target image $\mathbf{x}^{ut}_i \in \mathcal{D}_{ut}$, we calculate two similarity distributions $P^t_i$ and $P^{t\rightarrow s}_i$ as:
\begin{equation}
        P^t_{i,j}=\frac{\exp(\mu^t_j\cdot F(\mathbf{x}^{ut}_i)/t)}{\sum_{k=1}^c\exp(\mu^t_k\cdot F(\mathbf{x}^{ut}_i)/t)}
\label{eq:similarity-1}
\end{equation}
\begin{equation}
        P^{t\rightarrow s}_{i,j}=\frac{\exp(\mu^s_j\cdot F(\mathbf{x}^{ut}_i)/t)}{\sum_{k=1}^c\exp(\mu^s_k\cdot F(\mathbf{x}^{ut}_i)/t)},
\label{eq:similarity-2}
\end{equation}
where $t$ is a temperature value. Similarly, we can calculate $P^s_i$ and $P^{s\rightarrow t}_i$ for each source image $\mathbf{x}^{s}_i \in \mathcal{D}_{ls} \cup \mathcal{D}_{us}$. Therefore, the $\mathcal{L}_{self}$ objective in the baseline model can be formulated as:
\begin{equation}
\begin{aligned}
    \mathcal{L}_{self}=\sum_{i=1}^{N_{ls}+N_{us}}(&\mathcal{L}_{CE}(P^{s}_i,c_s(i))+\mathcal{H}(P^{s\rightarrow t}_i)) \\
        + \sum_{i=1}^{N_{ut}}(&\mathcal{L}_{CE}(P^{t}_i,c_t(i))+\mathcal{H}(P^{t\rightarrow s}_i)),
\end{aligned}
\end{equation}
where $\mathcal{H}(\cdot)$ is the entropy metric and $c(\cdot)$ denotes the cluster index of the corresponding sample.

\subsection{Details for Model Training}

\textbf{Backbone Choices.} 
We use ResNet-50~\cite{DBLP:conf/cvpr/HeZRS16} pretrained on ImageNet~\cite{russakovsky2015imagenet} as our backbone $F(\cdot)$ when validating our method on the Office-31, the Office-Home and the VisDA-C datasets following~\cite{DBLP:journals/corr/abs-2003-08264,DBLP:conf/cvpr/YueZZ0DKS21}. On the DomainNet dataset, we use ResNet-101 pretrained on ImageNet following PCS~\cite{DBLP:conf/cvpr/YueZZ0DKS21} for a fair comparison. In the meantime, we use ResNet-50 with pre-trained generalization weights provided by BrAD~\cite{harary2022unsupervised} for fair comparisons with FUDA results in BrAD.

\textbf{Model Structure Details.}
For fair comparison with~\cite{DBLP:journals/corr/abs-2003-08264,DBLP:conf/cvpr/YueZZ0DKS21,harary2022unsupervised}, we replace the last fully-connected layer with a randomly initialized linear layer and set the output feature dimension as 512. We perform the L2-normalization on the output features before sending them to the classifier $\phi(\cdot)$.

\textbf{Hyper-parameter Choices.}
During model training, we use the SGD optimizer with a momentum of 0.9. As for the model learning rate, we choose 1e-2 for Office-31, Office-Home, DomainNet (comparing with PCS), 1e-3 for VisDA-C, and 1e-4 for DomainNet (comparing with BrAD). Throughout the training, we fix the batch size at 64. We simply set the values of $\lambda_{MI}$ and $\lambda_{self}$ as in~\cite{DBLP:conf/cvpr/YueZZ0DKS21}. For visual dispersal objectives, we empirically set $\alpha=0.75$, $\lambda_{\text{X-VD}}=1.0$, $\lambda_{\text{I-VD}}=0.1$, $r^{\text{X}}_{h}\in\{0.75,0.85\}$, $r^{\text{I}}_{E}=0.1$ and $r^{\text{I}}_{H}\in\{0.65,0.75\}$.

\textbf{Training Device Choices.}
We use one NVIDIA GeForce RTX 3090 GPU for training and evaluation. 

\subsection{Implementation on the VisDA-C dataset}
\label{sec:visda_efficient}
On the Office-31~\cite{saenko2010adapting}, the Office-Home~\cite{venkateswara2017deep}, and the DomainNet~\cite{peng2019moment} datasets, we follow~\cite{DBLP:conf/cvpr/YueZZ0DKS21} and utilize the k-means clustering for $\mathcal{L}_{self}$ when training the baseline model.
On the significantly bigger VisDA-C dataset, to improve the training efficiency, we substitute the time-consuming k-means clustering with the attention-based prototype generating strategy in~\cite{liang2021source} and conduct only source-to-target transfer learning in loss $\mathcal{L}_{self}$.

\textbf{Simplified $\mathcal{L}_{self}$ in the baseline model.}
As shown in \cref{tab:dataset}, the VisDA-C~\cite{peng2017visda} dataset is significantly bigger. Training with k-means clustering on the VisDA-C dataset leads to unacceptable time costs. In order to improve the training efficiency of our proposed C-VisDiT, we substitute the time-consuming k-means approach with an attention-based strategy similar to~\cite{liang2021source}. We also provide an easier form of $\mathcal{L}_{self}$ to further reduce the calculation complexity. 
Specifically, for each source sample $\mathbf{x}^{s}_i \in \mathcal{D}_{ls} \cup \mathcal{D}_{us}$, we construct a memory bank to store the source sample features and update it by momentum in each epoch in order to reduce the impact of training fluctuation.
The memory bank is denoted as $B_{s}=[\mathbf{f}^{s}_1,\mathbf{f}^{s}_2,\cdots,\mathbf{f}^{s}_{N_s}]$ where $\mathbf{f}^{s}_i$ is the feature of $\mathbf{x}^{s}_i$ inside the memory bank.
We obtain the semantic distribution $p(y|\mathbf{f}^{s}_i)=[p_1,p_2,\cdots,p_c]$ for each source sample using the feature stored inside the memory bank. 
We concatenate the semantic distributions for all source samples as $K_{s}=[p(y|\mathbf{f}^{s}_1)^T,p(y|\mathbf{f}^{s}_2)^T, \cdots, p(y|\mathbf{f}^{s}_{N_s})^T]$, where $N_s=N_{ls}+N_{us}$. 
We then construct the source class centers in an attention-based manner:
\begin{equation}
    [\hat{\mu}^{s}_1, \hat{\mu}^{s}_2, \cdots, \hat{\mu}^{s}_c]= B_{s}K_{s}^T.
\end{equation}
With the attention-based source class centers $\hat{\mu}^{s}_j$, we calculate the similarity distribution for each target sample $\hat{P}^{t\rightarrow s}_i$ following \cref{eq:similarity-2}. We formulate the simplified $\mathcal{L}_{self}$ objective as:
\begin{equation}
    \mathcal{L}_{self}=\sum_{i=1}^{N_{ut}}(\mathcal{H}(\hat{P}^{t\rightarrow s}_i)).
\end{equation}



\textbf{Other implementation details.}
It is important to choose a decent initialization for training on the VisDA-C dataset. We train the model backbone with labeled source samples in a fully-supervised manner and utilize the weights of the trained backbone as the initialization for the training. Besides, during the first training epoch, we freeze the classification head $\phi(\cdot)$ and train our model with only $\mathcal{L}_{cls} + \lambda_{MI} \cdot \mathcal{L}_{MI}$. 

\section{Detailed Results for each Adaptation Setting on the DomainNet Dataset}
\label{sec:domainnet}

\textbf{DomainNet}~\cite{peng2019moment} is a challenging large-scale domain adaptation benchmark featuring 126 object classes. We present the overall performance of our C-VisDiT model on the DomainNet dataset in Tab. 4, Page 6 of the article. Here we present the detailed performance for each adaptation setting in \cref{tab:domainnet}. 
According to \cref{tab:domainnet}, our C-VisDiT can achieve 1.2\%/2.0\% and 1.3\%/2.5\% accuracy gain (1-shot/3-shots) comparing with PCS~\cite{DBLP:conf/cvpr/YueZZ0DKS21} and BrAD~\cite{harary2022unsupervised}, respectively. Looking into each adaptation setting, our C-VisDiT can realize state-of-the-art results on 25 out of 28 settings. 
These results show that our proposed C-VisDiT establishes new state-of-the-art performance on the most challenging benchmark for FUDA, well demonstrating its superiority.

\begin{table}[t]
\centering
\caption{Adaptation accuracy (\%) comparison on 1-shot and 3-shots labeled source per class on the DomainNet dataset. }
\resizebox{1.0\columnwidth}{!}{
\begin{threeparttable}
\begin{tabular}{l|cccccccc}
\toprule
\multirow{2}{*}{Method} & \multicolumn{8}{c}{DomainNet: Target Acc.} \\ \cmidrule{2-9} 
 & R$\rightarrow$C & R$\rightarrow$P & R$\rightarrow$S & P$\rightarrow$C & P$\rightarrow$R & C$\rightarrow$S & S$\rightarrow$P & Avg \\
\midrule
\multicolumn{9}{c}{\textbf{1-shot labeled source}} \\
\midrule
Source Only & 15.9 & 22.1 & 10.5 & 12.8 & 18.6 & 5.7 & 5.8 & 13.1 \\
MME~\cite{saito2019semi} & 13.8 & 29.2 & 9.7 & 16.0 & 26.0 & 13.4 & 14.4 & 17.5 \\
CDAN~\cite{long2018conditional}   & 16.0 & 25.7 & 12.9 & 12.6 & 19.5 & 7.2 & 8.0 & 14.6 \\
MDDIA~\cite{jiang2020implicit} & 18.0 & 30.6 & 15.9 & 15.4 & 27.4 & 9.3 & 10.2 & 18.1 \\
CDS~\cite{DBLP:journals/corr/abs-2003-08264} & 21.7 & 30.1 & 18.2 & 17.4 & 20.5 & 18.6 & 22.7 & 21.5 \\ \midrule
PCS~\cite{DBLP:conf/cvpr/YueZZ0DKS21} & 39.0 & 51.7 & \textbf{39.8} & 26.4 & 38.8 & \textbf{23.7} & 23.6 & 34.7 \\ 
C-VisDiT (Ours) & \textbf{39.1} & \textbf{52.2} & 38.1 & \textbf{27.6} & \textbf{43.8} & 23.3 & \textbf{27.1} & \textbf{35.9} \\
\midrule
BrAD~\cite{harary2022unsupervised} & 48.6 & 55.1 & 52.8 & 44.6 & 47.8 & 47.9 & 51.0 & 49.7 \\
+C-VisDiT (Ours) & \textbf{51.0} & \textbf{55.8} & \textbf{54.2} & \textbf{45.6} & \textbf{47.9} & \textbf{49.9} & \textbf{52.3} & \textbf{51.0} \\ 
\midrule
\multicolumn{9}{c}{\textbf{3-shot labeled source}} \\ \midrule
Source Only & 23.7 & 40.3 & 22.9 & 19.3 & 48.3 & 19.1 & 15.8 & 27.1 \\
MME~\cite{saito2019semi} & 22.8 & 46.5 & 14.5 & 25.1 & 50.0 & 20.1 & 24.9 & 29.1 \\
CDAN~\cite{long2018conditional} & 30.0 & 40.1 & 21.7 & 21.4 & 40.8 & 17.1 & 19.7 & 27.3 \\
MDDIA~\cite{jiang2020implicit} & 41.4 & 50.7 & 37.4 & 31.4 & 52.9 & 23.1 & 24.1 & 37.3 \\
CDS~\cite{DBLP:journals/corr/abs-2003-08264} & 44.5 & 52.2 & 40.9 & 40.0 & 47.2 & 33.0 & 40.1 & 42.5 \\ \midrule
PCS~\cite{DBLP:conf/cvpr/YueZZ0DKS21} & 45.2 & \textbf{59.1} & 41.9 & 41.0 & 66.6 & 31.9 & 37.4 & 46.1 \\
C-VisDiT (Ours) & \textbf{48.2} & 58.7 & \textbf{42.1} & \textbf{41.6} & \textbf{68.3} & \textbf{32.5} & \textbf{45.3} & \textbf{48.1} \\ 
\midrule
BrAD~\cite{harary2022unsupervised} & 60.6 & 62.8 & 61.6 & 56.6 & 63.6 & 59.8 & 61.0 & 60.9 \\
+C-VisDiT (Ours) & \textbf{64.0} & \textbf{65.0} & \textbf{63.9} & \textbf{60.6} & \textbf{65.6} & \textbf{61.6} & \textbf{63.1} & \textbf{63.4} \\
\bottomrule
\end{tabular}
\end{threeparttable}
}
\label{tab:domainnet}
\end{table}

\begin{table}[]
\centering
\caption{Performance contribution of each component on the Office-Home (3\% / 6\% labeled source), the VisDA-C (1\% labeled source), and the DomainNet (1-shot / 3-shots labeled source) datasets in terms of adaptation accuracy (\%).}
\resizebox{1.0\columnwidth}{!}{%
\begin{tabular}{c|cc|c|c|c}
\toprule
{Method} & {$\mathcal{L}_{\text{X-VD}}$} & {$\mathcal{L}_{\text{I-VD}}$} & Office-Home & VisDA-C & DomainNet \\
\midrule
Baseline & $\times$ & $\times$ & 60.0 / 63.0 & 78.9 & 48.0 / 61.3 \\
C-VisDiT-X & $\surd$ & $\times$ & 61.5 / 64.5 & 79.6 & 50.5 / 62.8 \\
C-VisDiT-I & $\times$ & $\surd$ & 60.8 / 63.9 & 80.0 & 50.4 / 63.0 \\
C-VisDiT & $\surd$ & $\surd$ & \textbf{62.3} / \textbf{65.4} & \textbf{80.5} & \textbf{51.0} / \textbf{63.4} \\
\bottomrule
\end{tabular}%
}
\label{tab:home_visda_ablation_main}
\end{table}


\begin{figure}[t]
  \centering
   \includegraphics[width=0.90\linewidth, trim={0 0 0cm 1.3cm}, clip]{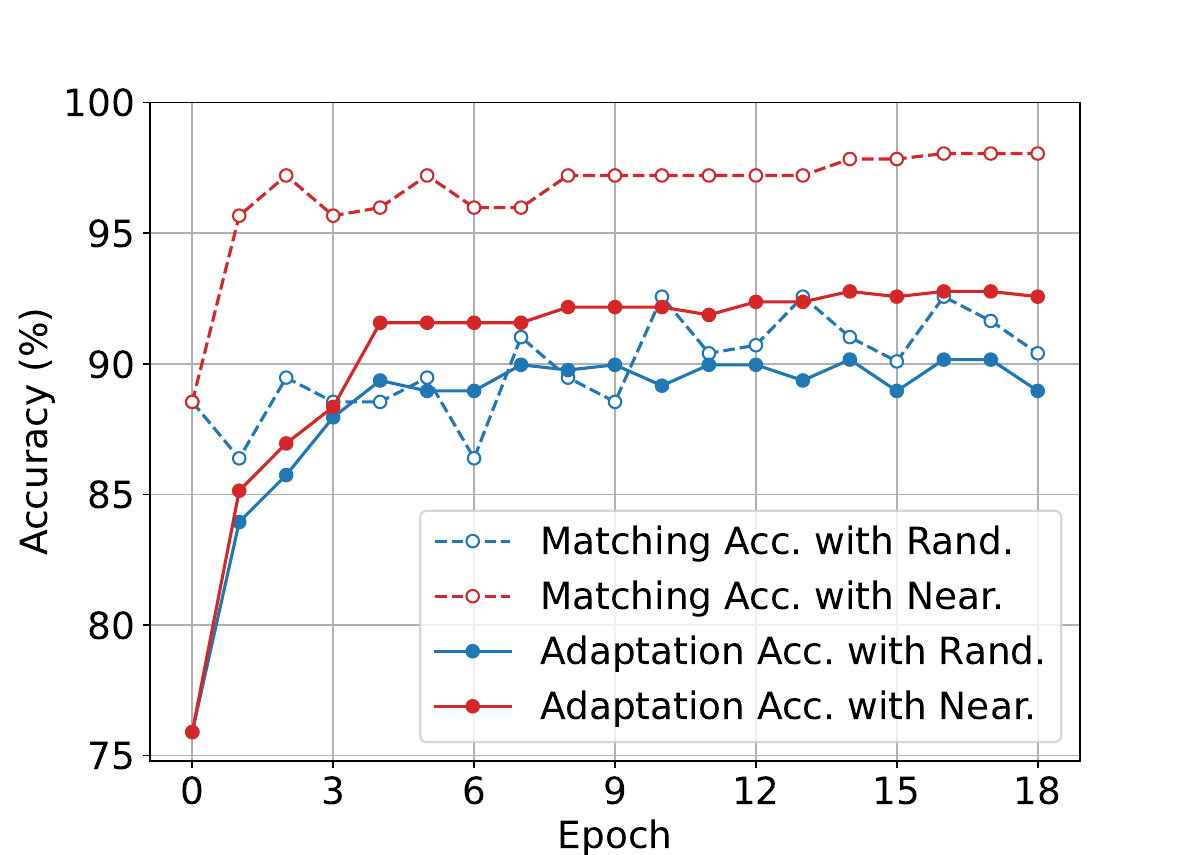}
   \caption{Analysis on the matching accuracy and the adaptation accuracy for both nearest source matching (``Near.'') and random source matching (``Rand.'') strategies in the W$\rightarrow$D (1-shot) setting on the Office-31 dataset.}
   \label{fig:matching_consistency}
\end{figure}

\begin{table}[]
\centering
\caption{Performance comparison of X-VD between different sample similarity measurement choices on the Office-31 dataset (\%). }
\vspace{-1mm}
\resizebox{0.7\columnwidth}{!}{%
\begin{threeparttable}
\begin{tabular}{l|c}
\toprule[1.0pt]
 Method & 1-shot / 3-shots \\ 
\midrule
 Cosine similarity & {79.1 / 84.7} \\
 Euclidean distance (Ours) & \textbf{79.2 / 84.9} \\
\bottomrule
\end{tabular}%
\end{threeparttable}
}
\label{tab:similarity_measurement}
\vspace{-2mm}
\end{table}

\begin{table}[]
\centering
\caption{Adaptation accuracy when $\mathcal{D}_{ut}^{E}$ is added to $\mathcal{T}^{H}_{ut}$ for C-VisDiT-I on the Office-31 dataset (\%).}
\resizebox{\columnwidth}{!}{%
\begin{tabular}{c|cc|c}
\toprule
 Method & $\mathcal{T}^{E}_{ut}$ & $\mathcal{T}^{H}_{ut}$ & 1-shot / 3-shots \\
\midrule
 Baseline & - & - & 77.6 / 83.8 \\ \midrule
 \multirow{2}{*}{C-VisDiT-I} & $\mathcal{D}_{ls}\cup \mathcal{D}_{ut}^{E}$ (\textbf{Ours}) & $\mathcal{D}_{ut}^{E}\cup \mathcal{D}_{ut}^{H}$ & 78.7 / \textbf{85.0} \\
 & $\mathcal{D}_{ls}\cup \mathcal{D}_{ut}^{E}$ (\textbf{Ours}) & $\mathcal{D}_{ut}^{H}$ (\textbf{Ours}) & \textbf{78.8} / \textbf{85.0} \\
\bottomrule
\end{tabular}%
}
\label{tab:office_compensate_IDVD}
\end{table}

\begin{table}[]
\centering
\caption{Comparison between C-VisDiT and PCS~\cite{DBLP:conf/cvpr/YueZZ0DKS21} in terms of the training efficiency on the VisDA-C dataset.}
\vspace{-1mm}
\resizebox{0.8\columnwidth}{!}{%
\begin{threeparttable}
\begin{tabular}{l|cc}
\toprule[1.0pt]
 & PCS~\cite{DBLP:conf/cvpr/YueZZ0DKS21} & C-VisDiT (Ours) \\ 
\midrule
 Training time (h) & 16.21 & 3.49 \\
 Target accuracy (\%)  & 79.0 & \textbf{80.5} \\
 Speedup & 1 & \textbf{4.64} \\
\bottomrule
\end{tabular}%
\end{threeparttable}
}
\label{tab:visda_time}
\vspace{-2mm}
\end{table}

\begin{table}[]
\centering
\caption{Performance comparison with other related works on the Office-31 dataset (\%).}
\vspace{-1mm}
\resizebox{0.55\columnwidth}{!}{%
\begin{threeparttable}
\begin{tabular}{l|c}
\toprule[1.0pt]
 Method & 1-shot / 3-shots \\ 
\midrule
 Baseline & 77.6 / 83.8 \\
 MixStyle~\cite{zhou2021domain} & 78.3 / 84.3 \\
 FlexMatch~\cite{zhang2021flexmatch} & 77.8 / 83.9 \\
 C-VisDiT (Ours) & \textbf{81.0 / 85.7} \\
\bottomrule
\end{tabular}%
\end{threeparttable}
}
\label{tab:more_related}
\vspace{-2mm}
\end{table}

\begin{table}[]
\centering
\caption{Adaptation accuracy with UDA methods using full source labels on the Office-31 dataset (\%).}
\resizebox{0.55\columnwidth}{!}{%
\begin{tabular}{c|cc}
\toprule
 Method & Origin & +C-VisDiT \\
\midrule
 DANN~\cite{ganin2015unsupervised} & 82.2 & \textbf{82.9} \\
 DSAN~\cite{zhu2020deep} & 88.4 & \textbf{88.8} \\
\bottomrule
\end{tabular}%
}
\label{tab:office_uda}
\end{table}

\section{Additional Analysis Experiment}


\textbf{Ablation Studies on the Office-Home, the VisDA-C, and the DomainNet datasets.} We present the result of ablation studies on the Office-Home~\cite{venkateswara2017deep}, the VisDA-C~\cite{peng2017visda}, and the DomainNet~\cite{peng2019moment} datasets in \cref{tab:home_visda_ablation_main}. We can see that both C-VisDiT-X and C-VisDiT-I can achieve better results compared to the baseline. Combining C-VisDiT-X and C-VisDiT-I, \ie, adding both X-VD and I-VD to the baseline model, results in the highest performance gain. These results, again, show that our X-VD and I-VD strategies are effective and complementary.

\textbf{Analysis on the nearest source matching in X-VD.}
In our proposed X-VD strategy, we conduct nearest source matching, where we match a given target sample to its nearest labeled source sample. To verify the effect of our nearest source matching, 
we analyze the matching accuracy, which reveals the consistency between the target sample and its neighboring labeled source sample. We compare our nearest source matching (``Near.'') with random source matching (``Rand.''), in which we match a given target sample to a random labeled source sample. As illustrated in \cref{fig:matching_consistency}, the employed nearest source matching can consistently obtain better matching accuracy than the random source matching (see dashed lines) during the training, which is in accordance with the comparison of the adaptation accuracy (see solid lines). 
These observations indicate that our nearest source matching can pull two samples with similar semantics but from different domains closer to each other, thus greatly boosting the adaptation.

\textbf{Comparison to other sample similarity measurement metrics.} In our proposed X-VD strategy, we utilize Euclidean distance to find similar samples across domains. To verify the effect of measurement choices, we replace Euclidean distance with the cosine similarity metric between feature vectors. According to \cref{tab:similarity_measurement}, both metrics yield similar performance. This suggests that the effectiveness of our method is not affected much by the sample similarity measurement metrics. 

\textbf{Necessity to train on easy target samples in I-VD.}
Fig. 4 in Page 8 of the article indicates that easy target samples suffer from slight accuracy loss during training. To show that it is unnecessary to train on easy target samples as a compensation for the accuracy loss, we further add easy target samples $\mathcal{D}^{E}_{ut}$ to the training hard sample set $\mathcal{T}^{H}_{ut}=\mathcal{D}^{E}_{ut} \cup \mathcal{D}^{H}_{ut}$. As shown in \cref{tab:office_compensate_IDVD}, adding $\mathcal{D}^{E}_{ut}$ to $\mathcal{T}^{H}_{ut}$ yields comparable results with the standard implementation $\mathcal{T}^{H}_{ut}=\mathcal{D}^{H}_{ut}$, indicating that it is not necessary to train on the easy target samples.

\begin{figure*}[htbp]
  \centering
\begin{subfigure}{.65\columnwidth}
    \centering
    \includegraphics[width=\linewidth, trim={0 0.5cm 0 0}, clip]{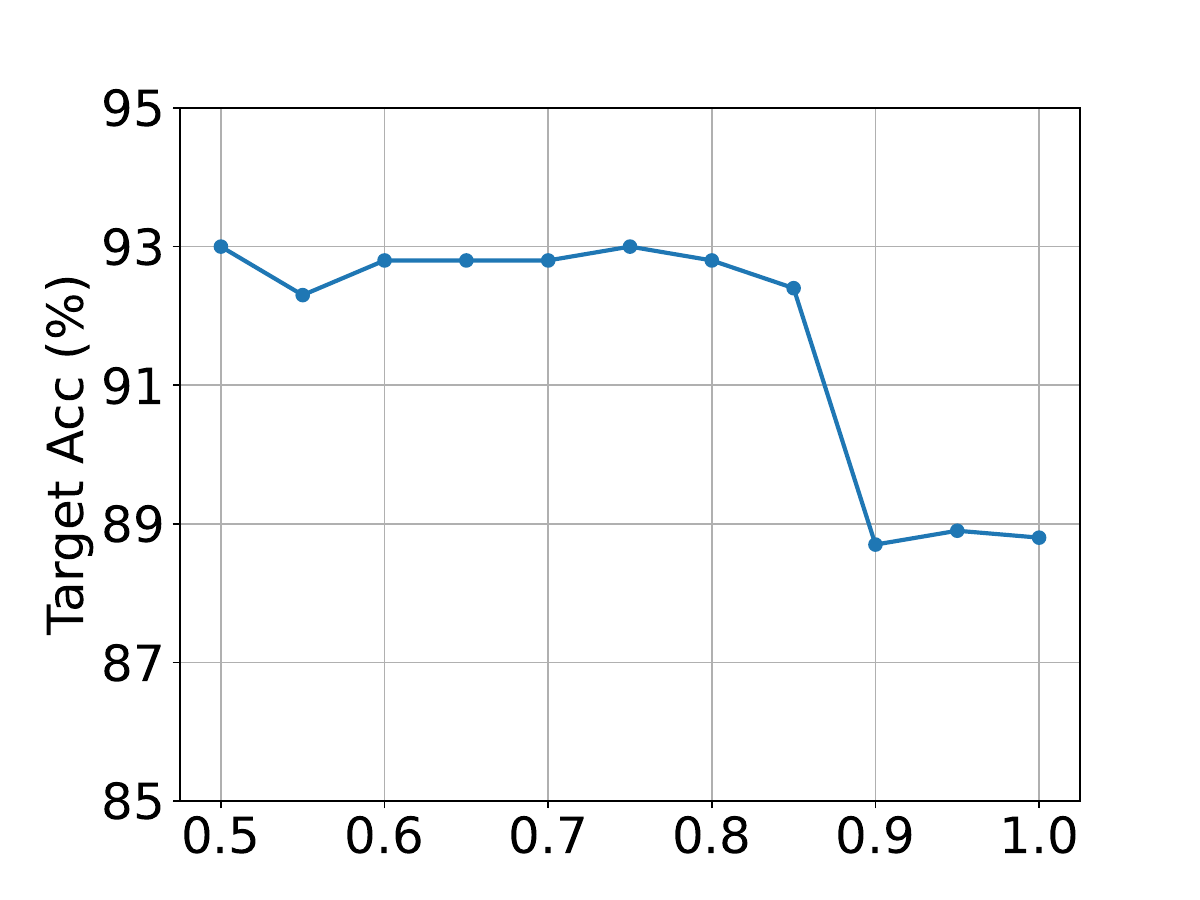}
    \centering
    \caption*{$r^{\text{X}}_h$ (C-VisDiT-X)}
    \label{fig:super_param-1}
\end{subfigure}%
\begin{subfigure}{.65\columnwidth}
    \centering
    \includegraphics[width=\linewidth, trim={0 0.5cm 0 0}, clip]{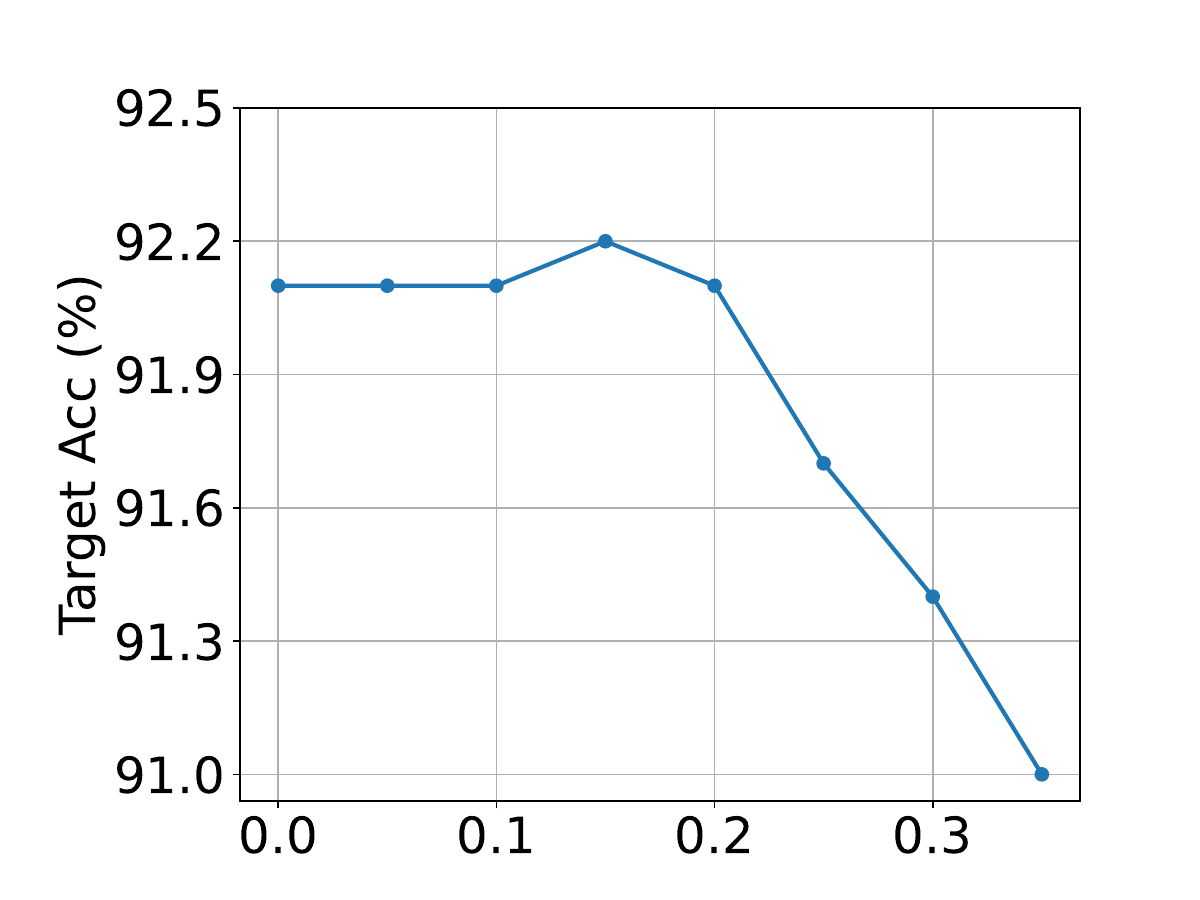}
    \caption*{\hspace*{0.5cm}$r^{\text{I}}_E$ (C-VisDiT-I)}
    \label{fig:super_param-2}
\end{subfigure}%
\begin{subfigure}{.65\columnwidth}
    \centering
    \includegraphics[width=\linewidth, trim={0 0.5cm 0 0}, clip]{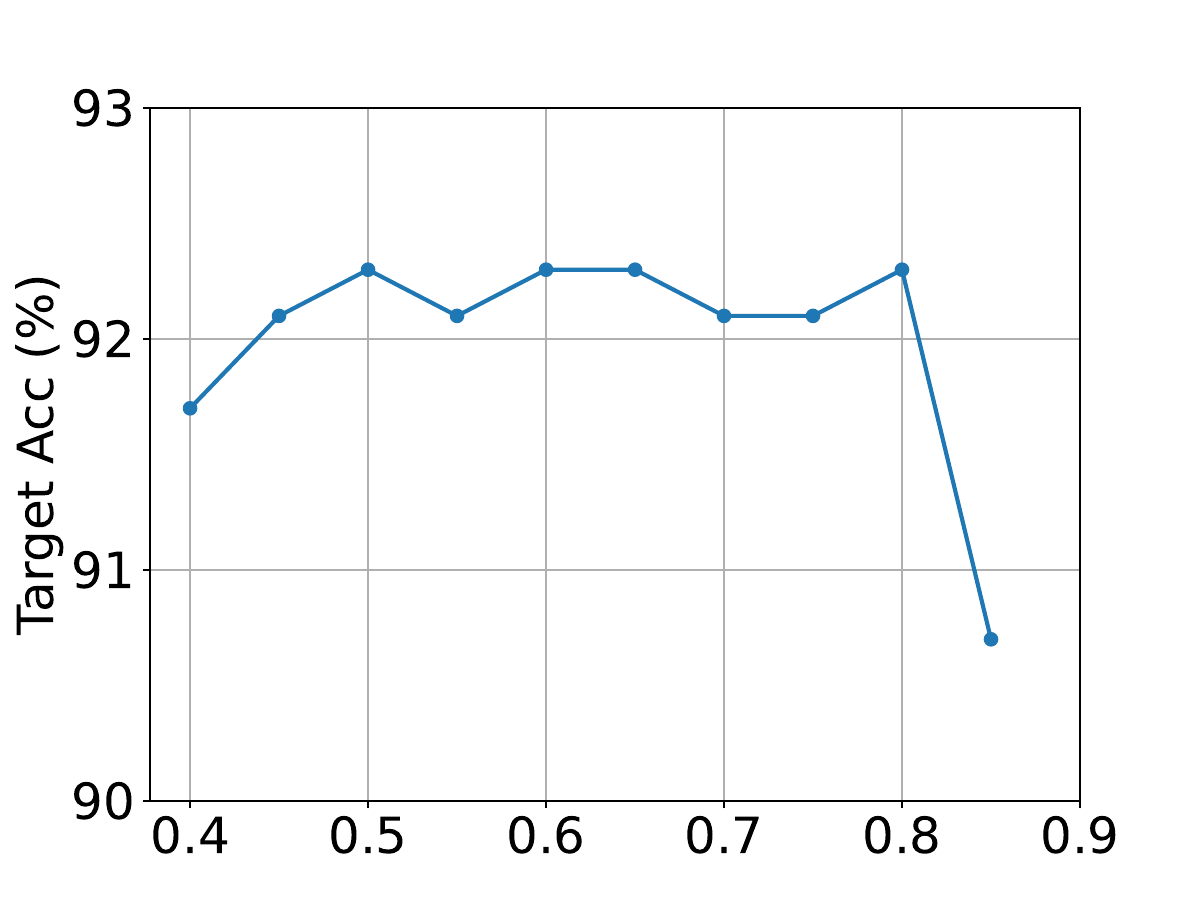}
    \caption*{\hspace*{0.3cm}$r^{\text{I}}_H$ (C-VisDiT-I)}
    \label{fig:super_param-3}
\end{subfigure}%
   \caption{Analysis of hyper-parameter $r^{\text{X}}_h$, $r^{\text{I}}_E$, and $r^{\text{I}}_H$ in D$\rightarrow$W (1-shot) on the Office-31 dataset. To reach promising performance, both $r^{\text{X}}_h$, $r^{\text{I}}_E$ and $r^{\text{I}}_H$ should be smaller than a value threshold (0.85 for $r^{\text{X}}_h$, 0.2 for $r^{\text{I}}_E$, and 0.8 for $r^{\text{I}}_H$). }
   \label{fig:ratio_ablation}
\end{figure*}

\begin{figure*}[htbp]
\centering
\begin{subfigure}{.65\columnwidth}
    \centering
    \includegraphics[width=\linewidth, trim={0 0.5cm 0 0}, clip]{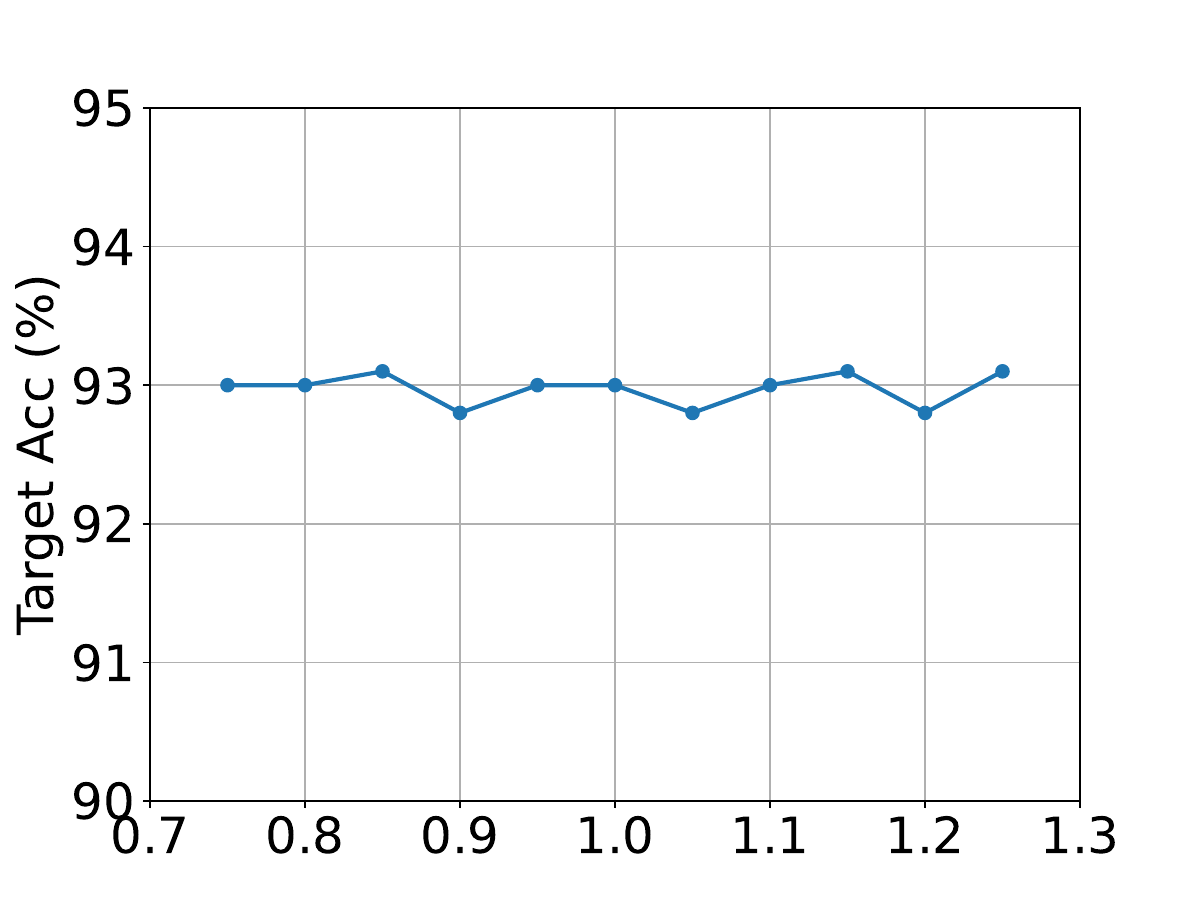}
    \caption*{$\lambda_{\text{X-VD}}$ (C-VisDiT-X)}
    \label{fig:super_param-4}
\end{subfigure}%
\begin{subfigure}{.65\columnwidth}
    \centering
    \includegraphics[width=\linewidth, trim={0 0.5cm 0 0}, clip]{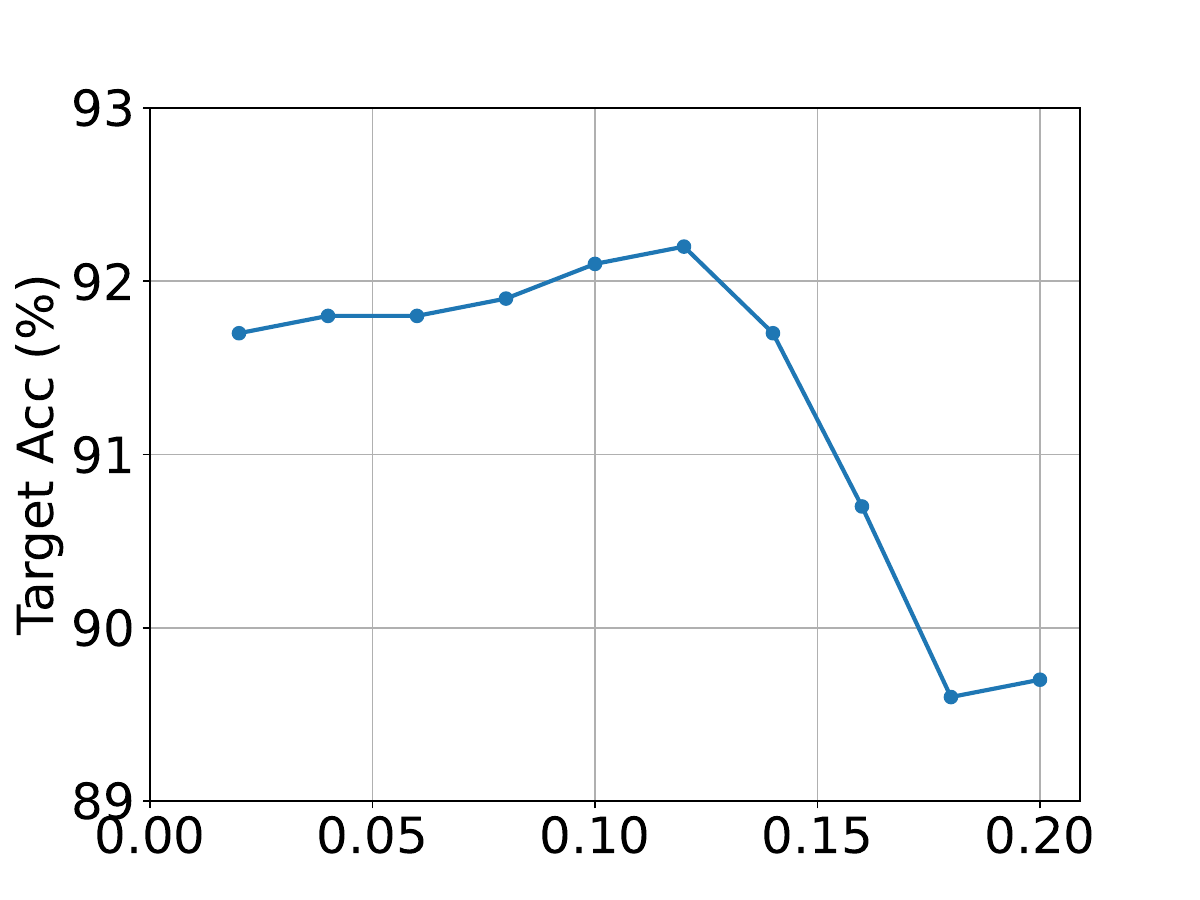}
    \caption*{$\lambda_{\text{I-VD}}$ (C-VisDiT-I)}
    \label{fig:super_param-5}
\end{subfigure}%
\begin{subfigure}{.65\columnwidth}
    \centering
    \includegraphics[width=\linewidth, trim={0 0.5cm 0 0}, clip]{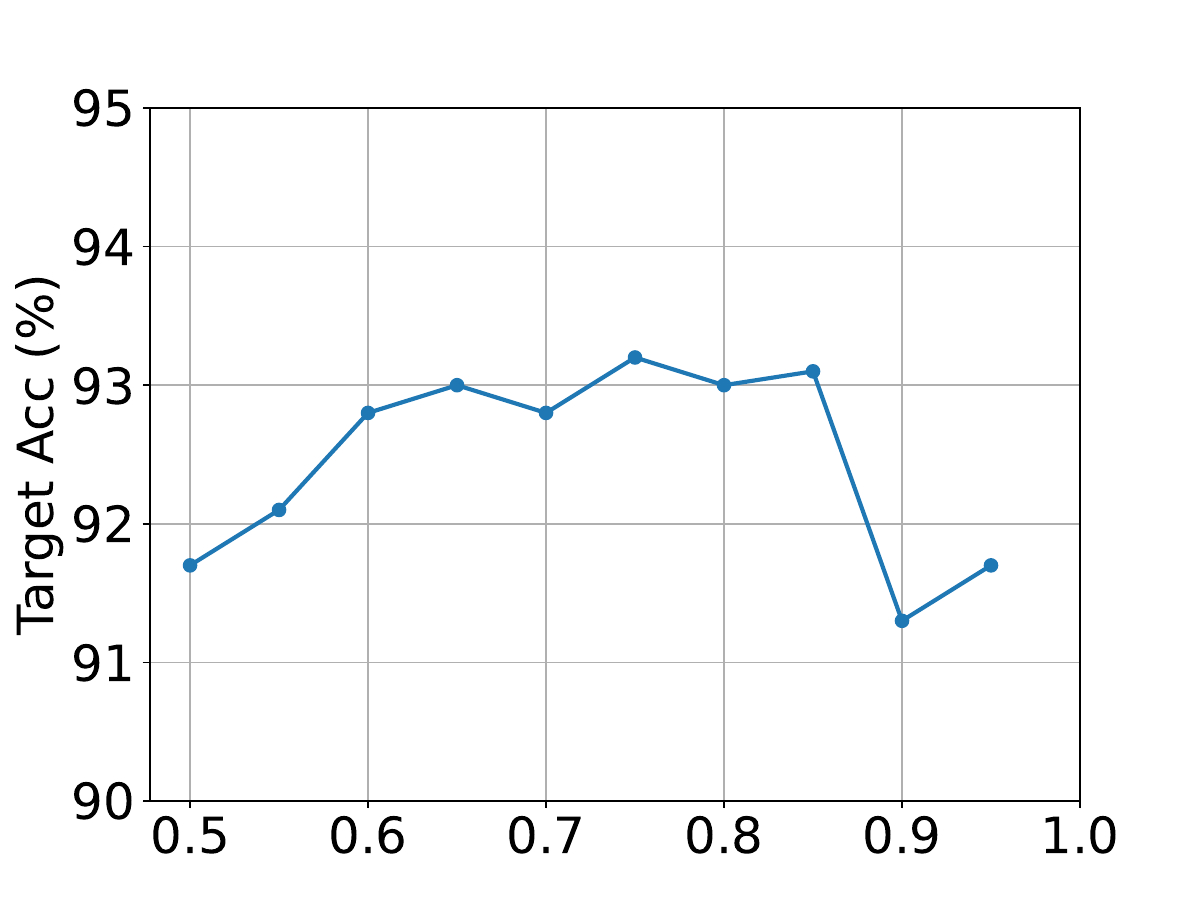}
    \caption*{$\alpha$ (C-VisDiT)}
    \label{fig:super_param-6}
\end{subfigure}%
   \caption{Analysis of hyper-parameter $\lambda_{\text{X-VD}}$, $\lambda_{\text{I-VD}}$, and $\alpha$ in D$\rightarrow$W (1-shot) on the Office-31 dataset. The value of $\lambda_{\text{X-VD}}$ hardly affects adaptation accuracy on the target domain, while the value of $\lambda_{\text{I-VD}}$ should be smaller than a value threshold (around 0.15). The value of $\alpha$ should be between a lower threshold (around 0.6) and an upper threshold (around 0.85). }
   \label{fig:lambda_alpha_ablation}
\end{figure*}

\textbf{Efficiency Comparison with PCS on VisDA-C.}
In \cref{sec:visda_efficient}, we claim that our simplified training pipeline on the VisDA-C dataset could effectively improve the training efficiency. To show the effect after simplifying the baseline model, we compare our simplified C-VisDiT with the existing state-of-the-art PCS~\cite{DBLP:conf/cvpr/YueZZ0DKS21} in terms of training efficiency. We train both methods on the VisDA-C dataset for 20 epochs from scratch, and report the training time and accuracy results. As indicated by \cref{tab:visda_time}, the proposed method can achieve significant speedup, \ie, 4.6 times compared with PCS, while enjoying a superior performance by 1.5\%. 
These results demonstrate that our simplifying strategies are especially effective on large datasets. 

\textbf{Comparison to other related works.} We provide performance comparisons with other related works in related fields, namely MixStyle~\cite{zhou2021domain} and FlexMatch~\cite{zhang2021flexmatch}. For MixStyle, we incorporate it into our baseline model using its published code. For FlexMatch, we implement it to FUDA by applying it to target samples based on the baseline model. The results are shown in \cref{tab:more_related}. Comparing to our C-VisDiT, both MixStyle and FlexMatch achieve inferior performance, and even worse than our C-VisDiT-X (79.2 / 84.9). Again, this demonstrates the superiority of our C-VisDiT, especially in the field of FUDA.

\section{Results on the UDA Problem}
To demonstrate the exclusive advantage of our C-VisDiT for FUDA, we further investigate the effect of C-VisDiT on the UDA problem using full source labels. 
We formally choose two remarkable UDA methods, \ie, DANN~\cite{ganin2015unsupervised} and DSAN~\cite{zhu2020deep}, as the baseline model, and equip them with our C-VisDiT. 
As shown in \cref{tab:office_uda}, our strategies still lead to performance gains, albeit limited (0.7\% on DANN and 0.4\% on DSAN). This can be attributed to that the knowledge is more confident in UDA due to its abundant supervision. 
As a result, the reliability of knowledge transfer can be guaranteed, leading to better adaptation on target samples.
Therefore, on the UDA problem, our confidence-based strategies cannot have significant effect as in FUDA. These results confirm that our confidence-based strategies have a distinctive edge for FUDA.

\section{Hyper-parameter Analysis}
\label{sec:hyperparameter}

\textbf{$r$ analysis.}
In our proposed C-VisDiT, we manually choose the ratios controlling the proportion of different confidence-level samples. The choice of $r$ hyper-parameters is correlated to the specific settings, as our model achieves different adaptation accuracy in different settings. Here we provide a detailed hyper-parameter analysis for the D$\rightarrow$W (1-shot) setting on the Office-31 dataset. We investigate the effect of $r^{\text{X}}_h$, $r^{\text{I}}_E$, and $r^{\text{I}}_H$ via adaptation accuracy. As shown in \cref{fig:ratio_ablation}, the adaptation performance basically remains stable in a wide range of $r$ values in the D$\rightarrow$W (1-shot) setting, \eg, $r^{\text{X}}_h\leq0.85$, $r^{\text{I}}_E\leq0.2$, and $r^{\text{I}}_H\leq0.8$. As a result, we choose $r^{\text{X}}_h=0.75$, $r^{\text{I}}_E=0.1$ and $r^{\text{I}}_H\leq0.65$ for the D$\rightarrow$W (1-shot) setting. For other settings, we empirically choose $r^{\text{X}}_{h}\in\{0.75,0.85\}$, $r^{\text{I}}_{E}=0.1$ and $r^{\text{I}}_{H}\in\{0.65,0.75\}$.


\begin{table}[]
\centering
\caption{Performance comparison of different hyper-paramameter $\beta$ choices on the Office-31 dataset (\%).}
\vspace{-1mm}
\resizebox{1.0\columnwidth}{!}{%
\begin{threeparttable}
\begin{tabular}{l|ccc}
\toprule[1.0pt]
 Method & $\beta>0.5$ & random $\beta$ & $\beta<0.5$ \\ 
\midrule
 C-VisDiT-X & \textbf{79.2\%/84.9\%}(Ours) & 78.8\%/84.8\% & 78.5\%/84.3\% \\
 C-VisDiT-I & 78.4\%/84.6\% & 78.6\%/84.7\% & \textbf{78.8\%/85.0\%}(Ours) \\
\bottomrule
\end{tabular}%
\end{threeparttable}
}
\label{tab:beta}
\vspace{-2mm}
\end{table}

\textbf{$\lambda_{\text{X-VD}}$ and $\lambda_{\text{I-VD}}$ analysis.}
We investigate the effect of $\lambda_{\text{X-VD}}$ and $\lambda_{\text{I-VD}}$ via adaptation accuracy in the D$\rightarrow$W (1-shot) setting on the Office-31 dataset. The results are shown in \cref{fig:lambda_alpha_ablation}. While the adaptation performance is not sensitive to $\lambda_{\text{X-VD}}$, it suffers from bigger values of $\lambda_{\text{I-VD}}$ in the D$\rightarrow$W (1-shot) setting. We empirically choose $\lambda_{\text{X-VD}}=1.0$ and $\lambda_{\text{X-VD}}=0.1$ in other settings for model evaluation experiments.


\textbf{$\alpha$ analysis.}
We investigate the effect of $\alpha$ utilized to sample $\beta$ in visual dispersal strategies, where $\beta \sim \text{Beta}(\alpha,\alpha)$. Similarly, we conduct experiments in the D$\rightarrow$W (1-shot) setting on the Office-31 dataset. The results are shown in \cref{fig:lambda_alpha_ablation}. Empirically, we choose $\alpha=0.75$ for the proposed C-VisDiT method in all settings.

\textbf{$\beta$ analysis.}
In Equation 10 and 14, Page 4 and 5 of the article, we use hyper-parameter $\beta$ to control the sample importance inside hybrid samples. In X-VD, we ensure that $\beta>0.5$ to put more importance on target samples to be learned. In I-VD, we guarantee that $\beta<0.5$ to put more importance on har target samples to be learned. As shown in \cref{tab:beta}, our choice of $\beta$ yields the best performance, implying that it is beneficial towards both X-VD and I-VD strategies.

\section{Image Retrieval Results}
\begin{figure*}[htbp]
  \centering
  \includegraphics[width=\linewidth, trim={0 0 0 0}, clip]{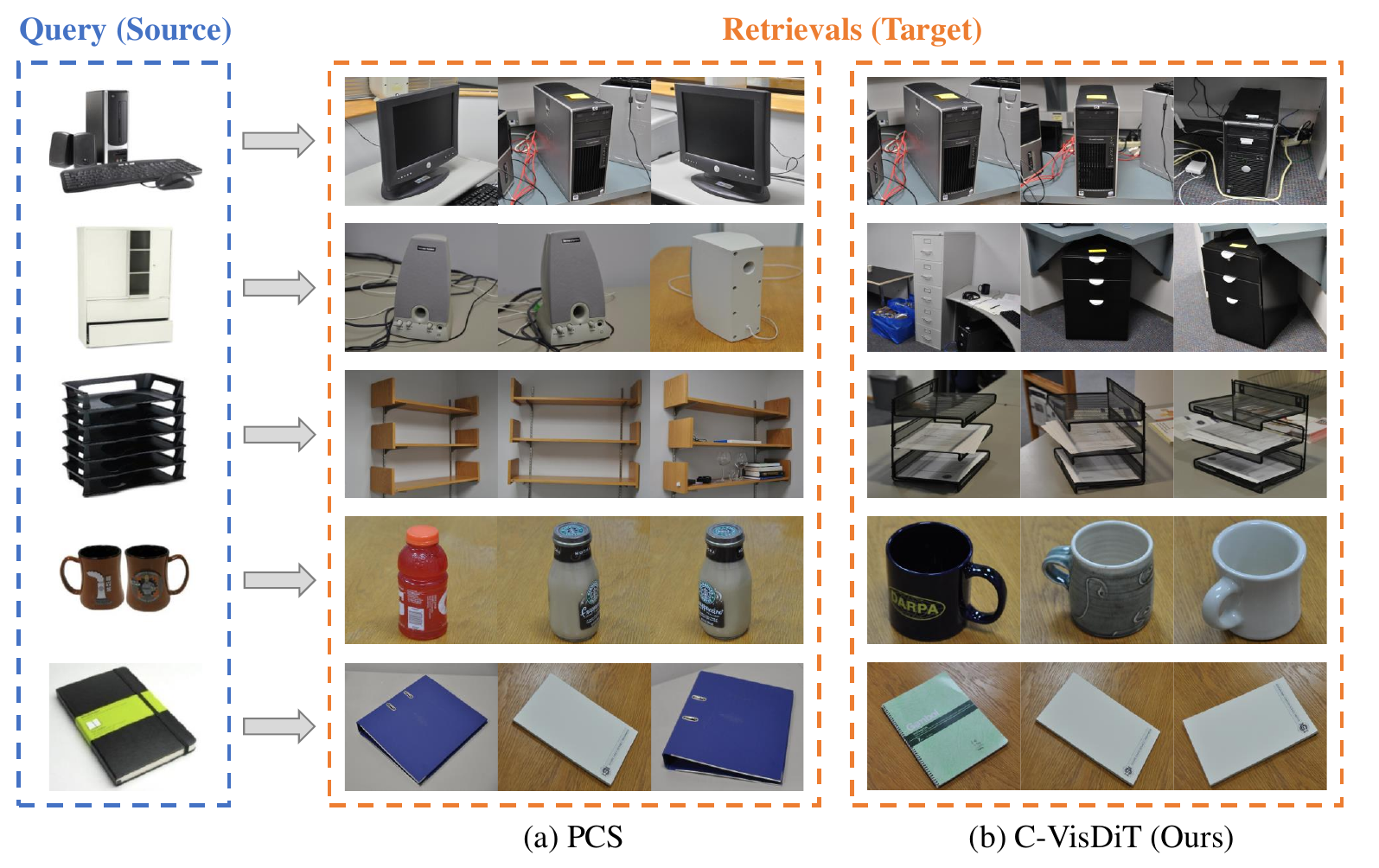}
  \caption{Image retrieval examples of the closest target domain samples given an unlabeled sample from the source domain, using PCS~\cite{DBLP:conf/cvpr/YueZZ0DKS21} (a) and our proposed C-VisDiT (b). The query images (from top to bottom) belong to the following categories: desktop computer, file cabinet, letter tray, mug, and paper notebook. PCS features tend to match images with similar visual patterns and textures. (Row 1: PCS matches monitors to the desktop computer query. Row 2: PCS matches speakers to the file cabinet query. Row 3: PCS matches bookcases to the letter tray query. Row 4: PCS matches bottles to the mug query. Row 5: PCS matches ring binders to the paper notebook query.) As a comparison, our C-VisDiT correctly matches images with similar semantics.}
  \label{fig:retrieval}
\end{figure*}

To qualitatively present that our proposed C-VisDiT can align semantically similar images across domains, we analyze our proposed C-VisDiT and the existing state-of-the-art, PCS~\cite{DBLP:conf/cvpr/YueZZ0DKS21}, via image retrieval. Given an unlabeled source sample $\mathbf{x}^{us}_i$, we find the three closest target domain samples measured by Euclidean distance in the feature space. As shown in \cref{fig:retrieval}, some features trained with PCS are aligned via visual textures and patterns instead of actual image semantics. For example, PCS tends to match letter trays (row 3) with bookcases, as both objects have similar layer structures. As a comparison, our proposed C-VisDiT aligns features that are semantically similar across domains, providing correct retrieval results for unlabeled source samples.


\clearpage



%% file: egpaper_final.bbl
\begin{thebibliography}{10}\itemsep=-1pt

\bibitem{abbet2021selfrule}
Christian Abbet, Linda Studer, Andreas Fischer, Heather Dawson, Inti Zlobec, Behzad Bozorgtabar, and Jean-Philippe Thiran.
\newblock Self-rule to adapt: Learning generalized features from sparsely-labeled data using unsupervised domain adaptation for colorectal cancer tissue phenotyping.
\newblock In {\em Medical Imaging with Deep Learning}, 2021.

\bibitem{DBLP:journals/pami/BadrinarayananK17}
Vijay Badrinarayanan, Alex Kendall, and Roberto Cipolla.
\newblock Segnet: {A} deep convolutional encoder-decoder architecture for image segmentation.
\newblock {\em {IEEE} Trans. Pattern Anal. Mach. Intell.}, 39(12):2481--2495, 2017.

\bibitem{baldock2021deep}
Robert Baldock, Hartmut Maennel, and Behnam Neyshabur.
\newblock Deep learning through the lens of example difficulty.
\newblock {\em Advances in Neural Information Processing Systems}, 34:10876--10889, 2021.

\bibitem{bousmalis2018using}
Konstantinos Bousmalis, Alex Irpan, Paul Wohlhart, Yunfei Bai, Matthew Kelcey, Mrinal Kalakrishnan, Laura Downs, Julian Ibarz, Peter Pastor, Kurt Konolige, et~al.
\newblock Using simulation and domain adaptation to improve efficiency of deep robotic grasping.
\newblock In {\em 2018 IEEE international conference on robotics and automation (ICRA)}, pages 4243--4250. IEEE, 2018.

\bibitem{bousmalis2016domain}
Konstantinos Bousmalis, George Trigeorgis, Nathan Silberman, Dilip Krishnan, and Dumitru Erhan.
\newblock Domain separation networks.
\newblock {\em Advances in Neural Information Processing Systems}, 29, 2016.

\bibitem{DBLP:conf/eccv/CarionMSUKZ20}
Nicolas Carion, Francisco Massa, Gabriel Synnaeve, Nicolas Usunier, Alexander Kirillov, and Sergey Zagoruyko.
\newblock End-to-end object detection with transformers.
\newblock In {\em European Conference on Computer Vision}, pages 213--229, 2020.

\bibitem{DBLP:journals/aei/ChenT21}
Ssu{-}Han Chen and Chia{-}Chun Tsai.
\newblock {SMD} {LED} chips defect detection using a yolov3-dense model.
\newblock {\em Adv. Eng. Informatics}, 47:101255, 2021.

\bibitem{ding2022proxymix}
Yuhe Ding, Lijun Sheng, Jian Liang, Aihua Zheng, and Ran He.
\newblock Proxymix: Proxy-based mixup training with label refinery for source-free domain adaptation.
\newblock {\em arXiv preprint arXiv:2205.14566}, 2022.

\bibitem{ganin2015unsupervised}
Yaroslav Ganin and Victor Lempitsky.
\newblock Unsupervised domain adaptation by backpropagation.
\newblock In {\em International conference on machine learning}, pages 1180--1189. PMLR, 2015.

\bibitem{DBLP:journals/jmlr/GaninUAGLLML16}
Yaroslav Ganin, Evgeniya Ustinova, Hana Ajakan, Pascal Germain, Hugo Larochelle, Fran{\c{c}}ois Laviolette, Mario Marchand, and Victor~S. Lempitsky.
\newblock Domain-adversarial training of neural networks.
\newblock {\em J. Mach. Learn. Res.}, 17:59:1--59:35, 2016.

\bibitem{ghifary2016deep}
Muhammad Ghifary, W~Bastiaan Kleijn, Mengjie Zhang, David Balduzzi, and Wen Li.
\newblock Deep reconstruction-classification networks for unsupervised domain adaptation.
\newblock In {\em European Conference on Computer Vision}, pages 597--613. Springer, 2016.

\bibitem{gretton2012kernel}
Arthur Gretton, Karsten~M Borgwardt, Malte~J Rasch, Bernhard Sch{\"o}lkopf, and Alexander Smola.
\newblock A kernel two-sample test.
\newblock {\em The Journal of Machine Learning Research}, 13:723--773, 2012.

\bibitem{harary2022unsupervised}
Sivan Harary, Eli Schwartz, Assaf Arbelle, Peter Staar, Shady Abu-Hussein, Elad Amrani, Roei Herzig, Amit Alfassy, Raja Giryes, Hilde Kuehne, et~al.
\newblock Unsupervised domain generalization by learning a bridge across domains.
\newblock In {\em Computer Vision and Pattern Recognition}, pages 5280--5290, 2022.

\bibitem{DBLP:conf/cvpr/HeZRS16}
Kaiming He, Xiangyu Zhang, Shaoqing Ren, and Jian Sun.
\newblock Deep residual learning for image recognition.
\newblock In {\em Computer Vision and Pattern Recognition}, pages 770--778, 2016.

\bibitem{hoffman2018cycada}
Judy Hoffman, Eric Tzeng, Taesung Park, Jun-Yan Zhu, Phillip Isola, Kate Saenko, Alexei Efros, and Trevor Darrell.
\newblock Cycada: Cycle-consistent adversarial domain adaptation.
\newblock In {\em International conference on machine learning}, pages 1989--1998, 2018.

\bibitem{huang2022small}
Haixin Huang, Xueduo Tang, Feng Wen, and Xin Jin.
\newblock Small object detection method with shallow feature fusion network for chip surface defect detection.
\newblock {\em Scientific reports}, 12(1):1--9, 2022.

\bibitem{jiang2020implicit}
Xiang Jiang, Qicheng Lao, Stan Matwin, and Mohammad Havaei.
\newblock Implicit class-conditioned domain alignment for unsupervised domain adaptation.
\newblock In {\em International conference on machine learning}, pages 4816--4827, 2020.

\bibitem{DBLP:journals/corr/abs-2003-08264}
Donghyun Kim, Kuniaki Saito, Tae{-}Hyun Oh, Bryan~A. Plummer, Stan Sclaroff, and Kate Saenko.
\newblock Cross-domain self-supervised learning for domain adaptation with few source labels.
\newblock {\em CoRR}, abs/2003.08264, 2020.

\bibitem{kim2017learning}
Taeksoo Kim, Moonsu Cha, Hyunsoo Kim, Jung~Kwon Lee, and Jiwon Kim.
\newblock Learning to discover cross-domain relations with generative adversarial networks.
\newblock In {\em International conference on machine learning}, pages 1857--1865. PMLR, 2017.

\bibitem{li2018semantic}
Peilun Li, Xiaodan Liang, Daoyuan Jia, and Eric~P Xing.
\newblock Semantic-aware grad-gan for virtual-to-real urban scene adaption.
\newblock {\em arXiv preprint arXiv:1801.01726}, 2018.

\bibitem{liang2021source}
Jian Liang, Dapeng Hu, Yunbo Wang, Ran He, and Jiashi Feng.
\newblock Source data-absent unsupervised domain adaptation through hypothesis transfer and labeling transfer.
\newblock {\em {IEEE} Trans. Pattern Anal. Mach. Intell.}, 2021.

\bibitem{DBLP:journals/corr/LiuRB15}
Wei Liu, Andrew Rabinovich, and Alexander~C. Berg.
\newblock Parsenet: Looking wider to see better.
\newblock {\em CoRR}, abs/1506.04579, 2015.

\bibitem{DBLP:conf/iccv/LiuL00W0LG21}
Ze Liu, Yutong Lin, Yue Cao, Han Hu, Yixuan Wei, Zheng Zhang, Stephen Lin, and Baining Guo.
\newblock Swin transformer: Hierarchical vision transformer using shifted windows.
\newblock In {\em International Conference on Computer Vision}, pages 9992--10002, 2021.

\bibitem{DBLP:conf/cvpr/LongSD15}
Jonathan Long, Evan Shelhamer, and Trevor Darrell.
\newblock Fully convolutional networks for semantic segmentation.
\newblock In {\em Computer Vision and Pattern Recognition}, pages 3431--3440, 2015.

\bibitem{long2015learning}
Mingsheng Long, Yue Cao, Jianmin Wang, and Michael Jordan.
\newblock Learning transferable features with deep adaptation networks.
\newblock In {\em International conference on machine learning}, pages 97--105, 2015.

\bibitem{long2018conditional}
Mingsheng Long, Zhangjie Cao, Jianmin Wang, and Michael~I Jordan.
\newblock Conditional adversarial domain adaptation.
\newblock {\em Advances in Neural Information Processing Systems}, 31, 2018.

\bibitem{long2016unsupervised}
Mingsheng Long, Han Zhu, Jianmin Wang, and Michael~I Jordan.
\newblock Unsupervised domain adaptation with residual transfer networks.
\newblock {\em Advances in Neural Information Processing Systems}, 29, 2016.

\bibitem{peng2019moment}
Xingchao Peng, Qinxun Bai, Xide Xia, Zijun Huang, Kate Saenko, and Bo Wang.
\newblock Moment matching for multi-source domain adaptation.
\newblock In {\em International Conference on Computer Vision}, pages 1406--1415, 2019.

\bibitem{peng2017visda}
Xingchao Peng, Ben Usman, Neela Kaushik, Judy Hoffman, Dequan Wang, and Kate Saenko.
\newblock Visda: The visual domain adaptation challenge.
\newblock {\em arXiv preprint arXiv:1710.06924}, 2017.

\bibitem{perone2019unsupervised}
Christian~S Perone, Pedro Ballester, Rodrigo~C Barros, and Julien Cohen-Adad.
\newblock Unsupervised domain adaptation for medical imaging segmentation with self-ensembling.
\newblock {\em NeuroImage}, 194:1--11, 2019.

\bibitem{radenovic2016cnn}
Filip Radenovi{\'c}, Giorgos Tolias, and Ond{\v{r}}ej Chum.
\newblock Cnn image retrieval learns from bow: Unsupervised fine-tuning with hard examples.
\newblock In {\em European conference on computer vision}, pages 3--20. Springer, 2016.

\bibitem{rakshit2022open}
Sayan Rakshit, Hmrishav Bandyopadhyay, Piyush Bharambe, Sai~Nandan Desetti, Biplab Banerjee, Subhasis Chaudhuri, et~al.
\newblock Open-set domain adaptation under few source-domain labeled samples.
\newblock In {\em Computer Vision and Pattern Recognition}, pages 4029--4038, 2022.

\bibitem{DBLP:conf/nips/RenHGS15}
Shaoqing Ren, Kaiming He, Ross~B. Girshick, and Jian Sun.
\newblock Faster {R-CNN:} towards real-time object detection with region proposal networks.
\newblock In {\em Advances in Neural Information Processing Systems}, pages 91--99, 2015.

\bibitem{russakovsky2015imagenet}
Olga Russakovsky, Jia Deng, Hao Su, Jonathan Krause, Sanjeev Satheesh, Sean Ma, Zhiheng Huang, Andrej Karpathy, Aditya Khosla, Michael Bernstein, et~al.
\newblock Imagenet large scale visual recognition challenge.
\newblock {\em International journal of computer vision}, 115(3):211--252, 2015.

\bibitem{saenko2010adapting}
Kate Saenko, Brian Kulis, Mario Fritz, and Trevor Darrell.
\newblock Adapting visual category models to new domains.
\newblock In {\em European Conference on Computer Vision}, pages 213--226, 2010.

\bibitem{saito2019semi}
Kuniaki Saito, Donghyun Kim, Stan Sclaroff, Trevor Darrell, and Kate Saenko.
\newblock Semi-supervised domain adaptation via minimax entropy.
\newblock In {\em International Conference on Computer Vision}, pages 8050--8058, 2019.

\bibitem{DBLP:conf/cvpr/ShrivastavaGG16}
Abhinav Shrivastava, Abhinav Gupta, and Ross~B. Girshick.
\newblock Training region-based object detectors with online hard example mining.
\newblock In {\em Computer Vision and Pattern Recognition}, pages 761--769, 2016.

\bibitem{shrivastava2017learning}
Ashish Shrivastava, Tomas Pfister, Oncel Tuzel, Joshua Susskind, Wenda Wang, and Russell Webb.
\newblock Learning from simulated and unsupervised images through adversarial training.
\newblock In {\em Computer Vision and Pattern Recognition}, pages 2107--2116, 2017.

\bibitem{sun2019unsupervised}
Yu Sun, Eric Tzeng, Trevor Darrell, and Alexei~A Efros.
\newblock Unsupervised domain adaptation through self-supervision.
\newblock {\em arXiv preprint arXiv:1909.11825}, 2019.

\bibitem{torralba2011unbiased}
Antonio Torralba and Alexei~A Efros.
\newblock Unbiased look at dataset bias.
\newblock In {\em Computer Vision and Pattern Recognition}, pages 1521--1528. IEEE, 2011.

\bibitem{tzeng2017adversarial}
Eric Tzeng, Judy Hoffman, Kate Saenko, and Trevor Darrell.
\newblock Adversarial discriminative domain adaptation.
\newblock In {\em Computer Vision and Pattern Recognition}, pages 7167--7176, 2017.

\bibitem{van2008visualizing}
Laurens Van~der Maaten and Geoffrey Hinton.
\newblock Visualizing data using t-sne.
\newblock {\em Journal of machine learning research}, 9(11), 2008.

\bibitem{venkateswara2017deep}
Hemanth Venkateswara, Jose Eusebio, Shayok Chakraborty, and Sethuraman Panchanathan.
\newblock Deep hashing network for unsupervised domain adaptation.
\newblock In {\em Computer Vision and Pattern Recognition}, pages 5018--5027, 2017.

\bibitem{DBLP:conf/cvpr/WangZZW20}
Dong Wang, Yuan Zhang, Kexin Zhang, and Liwei Wang.
\newblock Focalmix: Semi-supervised learning for 3d medical image detection.
\newblock In {\em Computer Vision and Pattern Recognition}, pages 3950--3959, 2020.

\bibitem{wang2016cost}
Keze Wang, Dongyu Zhang, Ya Li, Ruimao Zhang, and Liang Lin.
\newblock Cost-effective active learning for deep image classification.
\newblock {\em IEEE Transactions on Circuits and Systems for Video Technology}, 27(12):2591--2600, 2016.

\bibitem{wilson2020survey}
Garrett Wilson and Diane~J Cook.
\newblock A survey of unsupervised deep domain adaptation.
\newblock {\em ACM Transactions on Intelligent Systems and Technology}, 11(5):51:1--51:46, 2020.

\bibitem{wu2020dual}
Yuan Wu, Diana Inkpen, and Ahmed El-Roby.
\newblock Dual mixup regularized learning for adversarial domain adaptation.
\newblock In {\em European Conference on Computer Vision}, pages 540--555. Springer, 2020.

\bibitem{xu2020adversarial}
Minghao Xu, Jian Zhang, Bingbing Ni, Teng Li, Chengjie Wang, Qi Tian, and Wenjun Zhang.
\newblock Adversarial domain adaptation with domain mixup.
\newblock In {\em Proceedings of the AAAI conference on artificial intelligence}, volume~34, pages 6502--6509, 2020.

\bibitem{xuan2020hard}
Hong Xuan, Abby Stylianou, Xiaotong Liu, and Robert Pless.
\newblock Hard negative examples are hard, but useful.
\newblock In {\em European Conference on Computer Vision}, pages 126--142. Springer, 2020.

\bibitem{yan2020improve}
Shen Yan, Huan Song, Nanxiang Li, Lincan Zou, and Liu Ren.
\newblock Improve unsupervised domain adaptation with mixup training.
\newblock {\em arXiv preprint arXiv:2001.00677}, 2020.

\bibitem{yi2017dualgan}
Zili Yi, Hao Zhang, Ping Tan, and Minglun Gong.
\newblock Dualgan: Unsupervised dual learning for image-to-image translation.
\newblock In {\em International Conference on Computer Vision}, pages 2849--2857, 2017.

\bibitem{DBLP:conf/cvpr/YueZZ0DKS21}
Xiangyu Yue, Zangwei Zheng, Shanghang Zhang, Yang Gao, Trevor Darrell, Kurt Keutzer, and Alberto~L. Sangiovanni{-}Vincentelli.
\newblock Prototypical cross-domain self-supervised learning for few-shot unsupervised domain adaptation.
\newblock In {\em Computer Vision and Pattern Recognition}, pages 13834--13844, 2021.

\bibitem{DBLP:conf/eccv/ZeilerF14}
Matthew~D. Zeiler and Rob Fergus.
\newblock Visualizing and understanding convolutional networks.
\newblock In {\em European Conference on Computer Vision}, pages 818--833, 2014.

\bibitem{zhang2021flexmatch}
Bowen Zhang, Yidong Wang, Wenxin Hou, Hao Wu, Jindong Wang, Manabu Okumura, and Takahiro Shinozaki.
\newblock Flexmatch: Boosting semi-supervised learning with curriculum pseudo labeling.
\newblock {\em Advances in Neural Information Processing Systems}, 34:18408--18419, 2021.

\bibitem{DBLP:conf/iclr/ZhangCDL18}
Hongyi Zhang, Moustapha Ciss{\'{e}}, Yann~N. Dauphin, and David Lopez{-}Paz.
\newblock mixup: Beyond empirical risk minimization.
\newblock In {\em International Conference on Learning Representations}, 2018.

\bibitem{liru2022pneumonia}
Yang Zhang, Liru Qiu, Yongkai Zhu, Long Wen, and Xiaoping Luo.
\newblock A new childhood pneumonia diagnosis method based on fine-grained convolutional neural network.
\newblock {\em Computer Modeling in Engineering \& Sciences}, 133:873--894, 01 2022.

\bibitem{zhao2022review}
Sicheng Zhao, Xiangyu Yue, Shanghang Zhang, Bo Li, Han Zhao, Bichen Wu, Ravi Krishna, Joseph~E Gonzalez, Alberto~L Sangiovanni-Vincentelli, Sanjit~A Seshia, et~al.
\newblock A review of single-source deep unsupervised visual domain adaptation.
\newblock {\em IEEE Transactions on Neural Networks and Learning Systems}, 33(2):473--493, 2022.

\bibitem{zhou2021domain}
Kaiyang Zhou, Yongxin Yang, Yu Qiao, and Tao Xiang.
\newblock Domain generalization with mixstyle.
\newblock {\em arXiv preprint arXiv:2104.02008}, 2021.

\bibitem{zhu2017unpaired}
Jun-Yan Zhu, Taesung Park, Phillip Isola, and Alexei~A Efros.
\newblock Unpaired image-to-image translation using cycle-consistent adversarial networks.
\newblock In {\em International Conference on Computer Vision}, pages 2223--2232, 2017.

\bibitem{zhu2020deep}
Yongchun Zhu, Fuzhen Zhuang, Jindong Wang, Guolin Ke, Jingwu Chen, Jiang Bian, Hui Xiong, and Qing He.
\newblock Deep subdomain adaptation network for image classification.
\newblock {\em IEEE Transactions on Neural Networks and Learning Systems}, 32(4):1713--1722, 2020.

\bibitem{zhuo2017deep}
Junbao Zhuo, Shuhui Wang, Weigang Zhang, and Qingming Huang.
\newblock Deep unsupervised convolutional domain adaptation.
\newblock In {\em Proceedings of the 25th ACM international conference on Multimedia}, pages 261--269, 2017.

\end{thebibliography}
